\documentclass{article}

\PassOptionsToPackage{numbers}{natbib}
\usepackage[preprint]{neurips_2025}

\usepackage[utf8]{inputenc} % allow utf-8 input
\usepackage[T1]{fontenc}    % use 8-bit T1 fonts
\usepackage{hyperref}       % hyperlinks
\usepackage{url}            % simple URL typesetting
\usepackage{booktabs}       % professional-quality tables
\usepackage{amsfonts}       % blackboard math symbols
\usepackage{nicefrac}       % compact symbols for 1/2, etc.
\usepackage{microtype}      % microtypography
\usepackage{xcolor}         % colors
\usepackage{verbatim}
\usepackage{amsmath, amssymb}
\usepackage{todonotes}
\usepackage{dsfont}
\usepackage{bbm}
\usepackage{multirow}
\usepackage{makecell}
\usepackage{authblk}
\usepackage{bm}
\usepackage{siunitx}
\usepackage{tablefootnote}
\usepackage{threeparttable}
\usepackage{algorithm}
\usepackage{algpseudocode}
\title{Ethics-Aware Safe Reinforcement Learning for Rare-Event Risk Control in Interactive Urban Driving}

\author{
  Dianzhao Li \textsuperscript{1,2}\thanks{Corresponding author} \hspace{1cm}  Ostap Okhrin\textsuperscript{1,2}\\
  \textsuperscript{1}Chair of Econometrics and Statistics, esp. in the Transport Sector,\\
  Technische Universität Dresden\\
  Dresden, Germany \\
  \textsuperscript{2}Center for Scalable Data Analytics and Artificial Intelligence\\ 
  (ScaDS.AI) Dresden/Leipzig, Germany.\\
  \texttt{\{dianzhao.li, ostap.okhrin\}@tu-dresden.de} \\
}

\begin{document}

\maketitle

\begin{abstract}

Autonomous vehicles hold great promise for reducing traffic fatalities and improving transportation efficiency, yet their widespread adoption hinges on embedding credible and transparent ethical reasoning into routine and emergency maneuvers, particularly to protect vulnerable road users (VRUs) such as pedestrians and cyclists. Here, we present a hierarchical Safe Reinforcement Learning (Safe RL) framework that augments standard driving objectives with ethics-aware cost signals. At the decision level, a Safe RL agent is trained using a composite ethical risk cost, combining collision probability and harm severity, to generate high-level motion targets. A dynamic, risk-sensitive Prioritized Experience Replay mechanism amplifies learning from rare but critical, high-risk events. At the execution level, polynomial path planning coupled with Proportional-Integral–Derivative (PID) and Stanley controllers translates these targets into smooth, feasible trajectories, ensuring both accuracy and comfort. We train and validate our approach on closed-loop simulation environments derived from large-scale, real-world traffic datasets encompassing diverse vehicles, cyclists, and pedestrians, and demonstrate that it outperforms baseline methods in reducing risk to others while maintaining ego performance and comfort. This work provides a reproducible benchmark for Safe RL with explicitly ethics-aware objectives in human-mixed traffic scenarios. Our results highlight the potential of combining formal control theory and data-driven learning to advance ethically accountable autonomy that explicitly protects those most at risk in urban traffic environments. Across two interactive benchmarks and five random seeds, our policy decreases conflict frequency by $25\sim45\%$ compared to matched task successes while maintaining comfort metrics within 5\%.

\end{abstract}

\section{Introduction}

Over the past several decades, both academic institutions and industry stakeholders have devoted significant effort to advancing autonomous driving (AD) technologies. By eliminating human error, autonomous vehicles (AVs) promise to reduce traffic accidents and enhance both social welfare and economic productivity. While fully functional Level 5 systems are still some decades away \citep{khan2022level}, the transitional period will involve extensive interaction between AVs and human-driven vehicles. This raises a pressing societal question: \textit{Can AVs make responsible decisions in emergencies and everyday interactions, not only to protect their occupants but also to protect those outside the vehicle} \citep{othman2021public}\textit{?} This question is especially urgent for vulnerable road users (VRUs) such as pedestrians and cyclists, who bear a disproportionately high burden of severe injury and death in collisions. According to the U.S. National Highway Traffic Safety Administration, 7,522 pedestrians were killed and more than 67,000 injured in 2022 \citep{naumann2025pedestrian}. Safeguarding VRUs should therefore be a primary objective of technical design, regulation, and public policy \citep{campolettano2024baseline}. Beyond emergency responses, broader ethical and policy concerns have also emerged, including legal liability, data privacy, and equitable access to mobility solutions \citep{wang2020ethical, lin2016ethics, bonnefon2016social}. Early debates often focused on abstract dilemmas such as the “trolley problem”. Empirical work such as the Moral Machine Experiment has demonstrated that ethical preferences vary widely across cultures \citep{awad2018moral}. However, focusing on such extreme dilemmas risks obscuring the everyday ethical trade-offs AVs face, such as yielding to pedestrians at crosswalks, balancing risk across multiple agents, and ensuring equitable safety for VRUs. These frequent, small-scale interactions cumulatively determine both public safety and social trust in autonomy \citep{evans2020ethical}. \par

Recent research has begun embedding ethical reasoning into AVs by combining formal control techniques with data-driven learning. For example, \citep{wang2020ethical2} introduces a lexicographic risk-minimization framework that ranks road users by priority, while \citep{geisslinger2023ethical} proposes an ethical trajectory planning approach in which candidate paths are evaluated using a set of principles such as risk minimization, worst-off protection, and equal treatment. These approaches leverage optimization and verification tools, offering interpretability and theoretical guarantees. At the same time, they often rely on pre-defined rankings, hand-tuned parameters, or fixed weighting schemes, which may not adapt well to heterogeneous, real-world traffic. In parallel, Reinforcement Learning (RL) has gained traction in AD research \citep{kiran2021deep, li2023modified}, drawing on its success in high-dimensional control problems \citep{mnih2015human, wurman2022outracing,li2025autonomous}. RL methods can adapt through data and explore trade-offs automatically. Still, most RL-based driving studies have focused on safety constraints or reward shaping, without explicitly operationalizing ethical reasoning across multiple road users. This gap limits their ability to address fairness and accountability in shared traffic environments. \par

To bridge this gap, we propose EthicAR, an ethics-aware Safe RL framework that integrates moral considerations as a continuous design principle rather than an afterthought for rare emergencies. EthicAR separates the objectives of AVs into two components: \textit{standard driving goals} (e.g., safety and efficiency) and \textit{ethics-aware costs} across surrounding traffic participants. The framework is organized into two hierarchical levels: a \textit{decision level}, where a Safe RL algorithm uses ethics-inspired costs to select high-level motion targets; and an \textit{execution level}, where classical controllers ensure smooth, feasible, and comfortable trajectory following. To better learn from rare but critical events, we introduce a dynamic, risk-sensitive Prioritized Experience Replay mechanism. We train and evaluate EthicAR in closed-loop simulation environments derived from large-scale, real-world traffic datasets with heterogeneous road users, enabling reproducible and realistic evaluation across complex scenarios. \par

The contributions of this paper are as follows:

\begin{enumerate}
    \item \textbf{Ethics-Aware Safe RL.} We design a composite ethical risk cost that integrates collision probability and harm severity, enabling Safe RL to balance ego safety with fairness to other road users.
    \item \textbf{Dynamic PER for Rare Events.} We introduce a risk-sensitive replay scheme that improves learning efficiency in rare but safety-critical interactions, validated through ablation.
    \item \textbf{Two-Level Control Architecture.} Our framework separates high-level decision making (Safe RL) from low-level execution (classical controllers), ensuring both principled reasoning and stable motion.
    \item \textbf{Comprehensive Evaluation.} We benchmark EthicAR against state-of-the-art baselines in simulation environments derived from real-world data, with stratified analysis across diverse traffic scenarios.
\end{enumerate}

\textbf{Limitations.} While this study shows the potential of ethics-aware Safe RL for AVs, several limitations remain. First, our evaluation relies on dataset-derived closed-loop simulations, where surrounding agents are only partially reactive; fully interactive multi-agent dynamics require further validation. Second, the ethical cost function depends on chosen weights and harm models, which may reflect implicit normative assumptions; systematic calibration or participatory preference elicitation remains future work. Third, our probabilistic risk model assumes independence across timesteps, which may underestimate compounding risk. Finally, we do not claim formal safety guarantees; rather, we aim to provide a reproducible benchmark and a step toward more accountable autonomous decision-making.

\begin{figure}
    \centering
    \includegraphics[width=\linewidth]{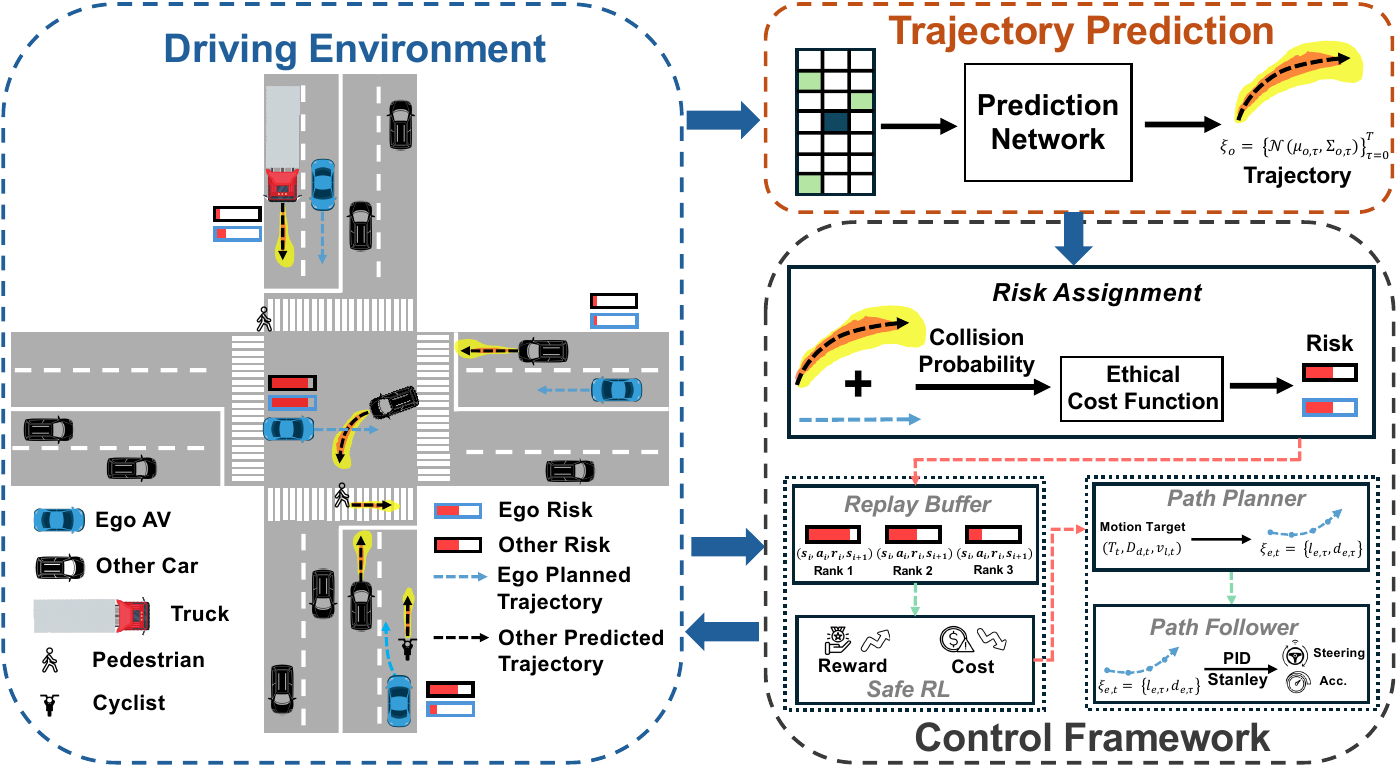}
    \caption{Overall diagram of the proposed data-driven EthicAR methodology. The hierarchical architecture comprises a decision level, where a Safe RL agent with an ethics‐aware cost function combining collision probability and harm severity, learns to output target motions. A dynamic PER emphasizes rare, high‐risk events during training. The execution level then converts these high‐level commands into smooth, feasible trajectories via a polynomial path planner and tracks them using PID/Stanley controllers, ensuring safe, comfortable maneuvers.}
    \label{fig:intro}
\end{figure}

\section{Background}
\label{sec:background}

\paragraph{Reinforcement learning.} RL is a foundational component of artificial intelligence, where an agent learns to make decisions through trial-and-error interactions with its environment \citep{Sutton2018}. This learning process is typically formalized as a Markov Decision Process (MDP), represented by the tuple $(\mathcal{S}, \mathcal{A}, \mathcal{P}, \mathcal{R}, \gamma )$. Here, $\mathcal{S}$ denotes the state space, capturing all possible configurations of the environment. $\mathcal{A}$ denotes the action space, describing the set of actions available to the agent. The transition probability function $\mathcal{P}: \mathcal{S} \times  \mathcal{A} \times  \mathcal{S} \rightarrow [0, 1] $ models the likelihood of transitioning from one state to another, given an action. The reward function 
$\mathcal{R}: \mathcal{S} \times \mathcal{A} \rightarrow \mathbb{R}$ assigns scalar feedback to each state-action pair. The discount factor $ \gamma \in [0,1)$ determines the relative importance of future rewards compared to immediate ones. Within this framework, the goal of the agent is to learn a policy  $\pi:\mathcal{S} \times \mathcal{A} \rightarrow [0,1]$, which maps states to a probability distribution over actions, so as to maximize the expected cumulative discounted return:

\begin{equation}
    \mathcal{J}(\pi) = \mathbb{E}_{\pi} \left [\sum_{t=0}^\infty  \gamma ^t \mathcal{R}(s_t, a_t, s_{t+1}) \right].
\end{equation}

To assess the effectiveness of a policy $\pi$, two value functions are commonly used. The state-value function, $V^\pi(s_t) = \mathbb{E}_\pi \left[ R_t \mid s_t \right]$, estimates the expected return when starting in state $s_t$ and following policy $\pi$ thereafter. The action-value function, $Q^\pi(s_t, a_t) = \mathbb{E}_\pi \left[ R_t \mid s_t, a_t \right]$, measures the expected return of taking action $a_t$ in state $s_t$, then following $\pi$. Based on how these value functions are utilized, RL methods are broadly categorized into two groups: \textit{value-based} and \textit{policy gradient} methods. In value-based approaches, an optimal policy $\pi^\star$ is derived by acting greedily to the optimal action-value function $Q^\star$. This optimal function satisfies the condition $Q^\star (s, a) = \max_\pi Q^\pi(s, a)$ for all $s\in \mathcal{S}$ and $a\in \mathcal{A}$. Conversely, policy gradient methods aim to directly optimize a parameterized policy $\pi_\theta$ by maximizing a performance objective, such as the expected return under $\pi_\theta$. A prominent subclass of these methods is the \textit{actor-critic} architecture, where the actor represents the policy and the critic estimates value functions to guide the updates of the actor \citep{haarnoja2018soft}.

The learning process can follow either an \textit{on-policy} or \textit{off-policy} paradigm. On-policy algorithms learn from data collected using the current policy, ensuring alignment between learning and exploration. In contrast, off-policy algorithms leverage experience collected from different policies, often achieving higher sample efficiency by enabling extensive data reuse. To further improve learning stability and efficiency, modern RL commonly uses experience replay. This involves storing past transitions in a replay buffer and sampling them during training. By disrupting the strong temporal correlations in sequential data, experience replay approximates the independent and identically distributed conditions assumed by gradient-based and temporal-difference methods. This contributes to more stable learning dynamics and reduced variance in updates.\par

\paragraph{Safe Reinforcement Learning.} Safe RL extends the RL paradigm by explicitly incorporating safety constraints into the learning process. This is formalized using a Constrained Markov Decision Process (CMDP), which augments the classic MDP tuple with an additional cost function $\mathcal{C}: \mathcal{S} \times \mathcal{A} \rightarrow \mathbb{R}$ and a predefined safety threshold $\eta$ \cite{altman2021constrained}. The cost function $\mathcal{C} (s, a) $ quantifies the risk or "unsafety" of taking action $a$ in state $s$. The expected cumulative cost under a given policy $\pi$ is defined as: $D({\pi}) = \mathbb{E} \left [\sum_{t=0}^\infty  \gamma ^t \mathcal{C}(s_t, a_t, s_{t+1}) \right]$. A policy is considered feasible if it satisfies the safety constraint, i.e., $D(\pi) \leq \eta$. The set of all such feasible policies is denoted as $\Pi_\mathcal{C} = \left\{ \pi \in \Pi | \mathcal{D}(\pi) \leq \eta \right\}$. The objective in Safe RL is to maximize the expected return while ensuring the cumulative cost remains within the allowed threshold. This leads to the following constrained optimization problem:

\begin{equation}
    \max_{\theta} \mathcal{J}(\pi_\theta), \text{s.t.} \ \pi_\theta \in \Pi_\mathcal{C}, 
\end{equation}
where $\pi_\theta$ is a parametrized policy. To solve such constrained problems, Lagrangian relaxation \citep{boyd2004convex} and projection-based policy updates \citep{achiam2017constrained} are commonly used. The Lagrangian method introduces a non-negative dual variable $\lambda \geq 0$ to convert the constrained problem into an unconstrained saddle-point formulation: 

\begin{equation}
    \max_\pi \min_{\lambda \geq 0} \left [ \mathcal{J}(\pi) - \lambda(D(\pi) - \eta)\right].
    \label{eq:safe rl}
\end{equation}

This approach enables simultaneous optimization of reward and adherence to safety constraints by adjusting the penalty term according to the degree of constraint violation. As a result, Safe RL provides a principled framework for training agents that must operate reliably in safety-critical environments.

\section{Methods}

\subsection{Problem Definition}
\label{sec:problem}

In this paper, we propose an AD framework that leverages RL to generate safe and socially compliant trajectories. The framework continuously processes real-time environmental data, including interactions with other vehicles, cyclists, and pedestrians, with more details in Section \ref{sec:obs and act}. Its goal is to balance efficiency, safety, and ethical behavior. To achieve this, we design a two-level control architecture with a decision level and an execution level. The decision level uses Safe RL to generate high-level motion targets. The execution level translates these targets into smooth and comfortable vehicle control. \par

The RL agent operates within a Frenet coordinate system, which describes motion relative to a reference path using two parameters: \textit{l}, the longitudinal distance along the path, and \textit{d}, the lateral offset from the path. This representation simplifies trajectory planning on curved roads by decoupling progress and deviation, providing an intuitive view of the vehicle's relationship to the road.  The RL agent should output a policy $\pi$, that maps observed states $s_t \in \mathcal{S}$ to actions $ a_t = (T_t, D_{d,t}, v_{l, t}) \in \mathcal{A}$, where $T_t$ denotes target planning time, $D_{d,t}$ defines target lateral displacement, and $v_{l, t}$ specifies target longitudinal speed along the reference path. The observation and action space, detailed in Section \ref{sec:obs and act}, are designed to ensure compatibility with dynamic road geometries. Using these outputs, the planner generates collision-free trajectories 
\[
\xi_{e} = [\textbf{p}_{e, \tau}]_{\tau = 1}^T, \quad \textbf{p}_{e, \tau} = [l_{e, \tau}, d_{e, \tau}], 
\]
where $\textbf{p}_{e, \tau}$ represents the target position in the Frenet coordinate system of the ego vehicle at each timestep $\tau$ within the target planning time. These trajectories are executed by the follower via low-level control, as described in Section \ref{sec:path generate and following}.\par

A key contribution of our framework is the integration of ethical risk minimization into the RL objective. Unlike traditional RL approaches, which focus solely on performance or safety, our method explicitly aims to reduce risks to \textbf{all road users}. Building on Safe RL principles introduced earlier, we formulate this as a constrained optimization problem. The agent seeks to maximize the expected return $\mathcal{J}_\pi$ while minimizing a cumulative ethical risk metric, defined as $D_{\text{ethics}} = \mathbb{E} \left [\sum_{t=0}^\infty  \gamma ^t \mathcal{C}_{\text{ethics}}(s_t, a_t, s_{t+1}) \right]$, where $\mathcal{C}_{\text{ethics}}$ quantifies potential risk to pedestrians, cyclists, and other road users. By incorporating this ethical cost, described in detail in Section~\ref{sec:cost}, directly into the constraint, our approach ensures that planned trajectories reflect not only a commitment to collision avoidance but also a broader sense of moral responsibility. For example, the agent learns to reduce unnecessary proximity to vulnerable individuals or to distribute unavoidable risk more equitably. This ethical integration sets our method apart from conventional AV planning techniques, which often treat safety and ethics as secondary considerations.

\subsection{LSTM based SACLag}

Soft Actor-Critic (SAC) algorithm combines actor-critic architecture with entropy regularization. It aims to maximize both cumulative rewards and policy entropy, encouraging exploration by favoring stochastic policies. SAC employs a temperature parameter $\alpha$ to balance reward maximization and entropy trade-offs. This temperature is automatically adjusted to maintain a target entropy level, avoiding manual tuning. The actor optimizes a stochastic policy, while the critic estimates Q-values using clipped double-Q networks to reduce overestimation bias \citep{haarnoja2018soft}. SAC with Lagrangian relaxation (SACLag) uses a Lagrangian multiplier to optimize while enforcing the cost constraint dynamically. It transforms the problem into a dual optimization: the primal variables (policy and Q-networks) maximize the reward-entropy objective, while the dual variable (Lagrangian multiplier) is adjusted via gradient ascent on the Lagrange loss. \par

As outlined in Section \ref{sec:problem}, the objective of the RL agent is to dynamically supply the path planner with relevant information for trajectory generation. However, individual states frequently lack the contextual depth needed to support optimal decision-making by the agent. In particular, historical information about the ego vehicle and surrounding traffic is important. To formally account for this, we model the problem as a partially observable Markov decision process (POMDP) \citep{kaelbling1998planning}, which extends the MDP tuple $(\mathcal{S}, \mathcal{A}, \mathcal{P}, \mathcal{R}, \gamma )$ with an observation space $\mathcal{O}$ and an observation function $\mathcal{Z} : \mathcal{S} \times \mathcal{A} \to \Delta(\mathcal{O})$, where $\Delta(\mathcal{O})$ denotes the probability distribution over $\mathcal{O}$. In a POMDP, the agent does not directly observe the true state $s_t \in \mathcal{S}$ but instead receives an observation $o_t \in \mathcal{O}$ generated according to $\mathcal{Z}(o_t \mid s_t, a_{t-1})$. To address this partial observability, we integrate Long Short-Term Memory (LSTM) networks \citep{hochreiter1997long} into the SACLag framework to deal with POMDPs. Each network in this framework incorporates LSTM layers, which process sequences of past time steps. This temporal reasoning capability enables the agent to retain and exploit historical information in combination with the current observation, thereby supporting more informed and consistent decision-making.

\subsection{Collision probabilities}
\label{sec:collision_prob}

To estimate the potential risk to other road users, we begin by computing the collision probabilities between the ego vehicle and its surrounding road users. Since the framework optimizes trajectories over varied time horizons, these probabilities are evaluated at each discrete timestep within that horizon. Given the planned trajectory of the ego vehicle, the next step is to predict the future trajectories of nearby road users. For surrounding vehicles, we use an LSTM-based neural network that incorporates convolutional social pooling and semantic road context, using bird’s-eye-view images as input \citep{geisslinger2021watch}. It provides at each timestep $\tau$ a mean position $\mu_{o, \tau}$ and a covariance matrix $\mathbf{\Sigma}_{o, \tau}$, defined over the longitudinal distance along the path and the lateral displacement. Under the Gaussian assumption, these parameters provide the probability distribution of the predicted positions of the agent. Additional implementation details are provided in \citep{geisslinger2021watch}. For other types of agents, such as pedestrians and cyclists, ground-truth trajectories are used with Gaussian noise added to capture uncertainty. The predicted trajectory of agent $o$ is thus represented as a sequence of Gaussian variables,
\[
\xi_{o,\tau} \sim \mathcal{N}(\mathbf{\mu}_{o,\tau}, \mathbf{\Sigma}_{o,\tau}), \quad \tau = 0, \dots, T, 
\quad \xi_o = \{ \xi_{o,\tau} \}_{\tau=0}^T.
\]

Unlike prior work that directly estimates collision probabilities, our method adopts a two-stage approach. In the first stage, we apply the Separating Axis Theorem (SAT) to check for potential overlaps between the ego vehicle and other agents, based on their mean trajectories \cite{gottschalk1996obbtree}. This check is performed at every timestep of the planning horizon. If an overlap is detected, we proceed to the second stage, where we compute a more precise proximity-based collision probability using the Mahalanobis distance. This two-stage approach improves efficiency by skipping expensive computations for clearly separated agents. \par

In the first stage, given the planned ego trajectories $\xi_e$ from our framework, we compute the overlap probabilities \(P_{\text{sat}}\) using the SAT. The SAT states that two convex shapes do not intersect if there exists at least one axis along which their projections are completely separated; conversely, if no such axis exists, the shapes must be colliding. This theorem is operationalized by determining potential separating axes, typically the normals of the edges of the convex bodies. In our context, these convex bodies are oriented bounding boxes (OBBs). Each shape is projected onto the candidate axes, and the intervals of projection are compared: a collision is confirmed if all projection intervals overlap, whereas the existence of even a single non-overlapping interval implies separation. This method has been widely adopted in the design of physics engines and simulation frameworks due to its computational efficiency and robustness in handling various orientations and configurations of convex objects. Let $\mathcal{U}^\tau = \left\{u_1^\tau,...,u_m^\tau \right\}$ be the set of separating axes for the OBB pair at time $\tau$. The overlap condition is:

\begin{equation}
\mathcal{O}_e^\tau \cap \mathcal{O}_o^\tau \neq \emptyset \Longleftrightarrow \forall u \in \mathcal{U}^\tau: \text{proj}_u (\mathcal{O}_e^\tau) \cap \text{proj}_u (\mathcal{O}_o^\tau) \neq \emptyset,
\end{equation}
with $\mathcal{O}_e^\tau$ and $\mathcal{O}_o^\tau$ indicates OBBs of ego and other traffic participants. The conditional overlap probability based on SAT is:

\begin{equation}
    P_{sat}^\tau =   \mathbbm{1} \left\{ \bigcap_{u \in \mathcal{U}^\tau} [\text{proj}_u (\mathcal{O}_e^\tau) \cap \text{proj}_u (\mathcal{O}_o^\tau) \neq \emptyset]  \right\} ,
\end{equation}
where $\mathbbm{1}$ is the indicator function.

If an overlap is detected, we compute the Mahalanobis distance between the ego vehicle and the predicted expected position of other traffic participants, taking into account uncertainties in the prediction. Assuming collisions at different timesteps are independent, the Mahalanobis distance between two traffic participants at each time $\tau$ is then expressed as:

\begin{equation}
    D_{m,\tau} = \sqrt{(\textbf{p}_{e, \tau} - \mathbf{\mu}_{o, \tau})^\top (\mathbf{\Sigma}_{o, \tau})^{-1} (\textbf{p}_{e, \tau} - \mathbf{\mu}_{o, \tau})}.
    \label{eq:maha dist}
\end{equation}

Assuming local Gaussianity \citep{geisslinger2021watch}, the squared Mahalanobis distance follows a chi-squared distribution with 2 degrees of freedom, i.e., $(D_{m,\tau})^2 \sim \chi^2(2)$. We convert the distance into a collision probability using the survival distribution function of the chi-squared distribution:

\begin{equation}
    P_{m}^\tau = 1 - F_{\chi^2} \left( (D_{m, \tau})^2; 2\right),
    \label{eq:col pro}
\end{equation}

where $F_{\chi^2}(\cdot; 2)$ is the chi-squared CDF with 2 degrees of freedom. Smaller Mahalanobis distances correspond to higher collision probabilities. By combining the two stages, we define the overall collision probability at each timestep $\tau$ as: 

\begin{equation}
    P^\tau =  P_{sat}^\tau \cdot P_{m}^\tau.
\end{equation}

This method enables efficient and accurate risk estimation. It avoids unnecessary computation when objects are clearly not in proximity, while still capturing nuanced spatial uncertainty when they are. Compared to Monte Carlo sampling, our approach offers substantial computational savings without sacrificing predictive reliability.

\subsection{Risk calculation}
\label{sec:risk_cal}

In this work, the risk $R_\tau$ associated with a traffic participant at time $\tau$ is defined as the expected collision risk with other traffic participants. This risk is calculated as the product of the collision probability ${P^\tau}$ and the estimated harm $H^\tau$, i.e., $R_\tau = P^\tau H^\tau$. While the collision probabilities are obtained from Section~\ref{sec:collision_prob}, we estimate the harm using empirical equations proposed in \citep{geisslinger2023ethical}. The harm $H$ between two traffic participants, $A$ and $B$, is modeled as:

\begin{equation}
    H = \frac{1}{1 + e^{c_0-c_1\Delta v -c_{a}}},
    \label{eq:harm model part 1}
\end{equation}

where $c_0$, $c_1$, and $c_{a}$ are empirical coefficients, and $\Delta v$ represents the effective collision speed of participant $A$, defined as:

\begin{equation}
    \Delta v_{A} =  {\frac{m_B}{m_A + m_B} \sqrt{v_A^2+v_B^2-2v_A v_B \cos \alpha}},
    \label{eq:harm model part 2}
\end{equation}
with $m$ and $v$ denoting the mass and velocity of the two road users $A$ and $B$, and $\alpha$ representing the angle between their velocities at the moment of collision. \par

Each potential collision is thus associated with a corresponding harm estimate. For a given planned trajectory of the ego vehicle, we compute a sequence of risk values $R_\tau$ across the planning horizon, each derived from the collision probability and associated harm at that timestep. To summarize this time-varying risk into a single scalar value, we take the maximum risk across the planning horizon $T$, expressed as: $R_{traj} = \max_{\tau\in[1, T]} R_\tau$. This yields a pairwise risk assessment for each interaction between the ego vehicle and another traffic participant. For each such pair, we assign a risk tuple ($R_{ego}, R_{obj}$), where $R_{ego}$ denotes the risk experienced by the ego vehicle, and $R_{obj}$ is the risk imposed on the other participant.

%Risks that originate from different collisions are assumed to be independent and thus calculated with equation (\ref{eq:risk}), where $S_R$ denotes the set of all other road users:
%\begin{equation}
%    R_{traj, ind} = 1 - \prod_{S_R}(1-{R_{traj}}).
%    \label{eq:risk}
%\end{equation}

%A combined risk from possible collisions with multiple road users only for the ego-vehicle in the case of the maximum acceptable risk principle for reasons of information asymmetry \cite{geisslinger2023ethical}.\par

\subsection{Cost function}
\label{sec:cost}

With the estimated risks for both the ego vehicle and surrounding road users, we design cost functions that promote fair and responsible risk distribution across all participants. These cost functions are integrated into the training process of the Safe RL agent, allowing the agent to learn behaviors that respect risk constraints while still optimizing the standard RL objectives. To evaluate the impact of different ethical considerations, we define two baseline modes: the \textit{ethical} mode and the \textit{selfish} mode. In the \textit{ethical} mode, the cost function accounts for the risks imposed on all traffic participants, encouraging decisions that are socially responsible. In contrast, the \textit{selfish} mode focuses solely on minimizing the risk to the ego vehicle, without regard for the safety of others.\par

For the \textit{ethical} mode, the cost function is expressed as:

\begin{equation}
    J_{\text{ethic}} = \omega_B J_B + \omega_E J_E + \omega_M J_M, 
    \label{eq:cost}
\end{equation}
where three main components are included, the Bayes principle $J_B$, the Equality principle $J_E$, and the Maximin principle $J_M$, The corresponding weights $\omega_B$, 
$\omega_E$, and $\omega_M$ determine the relative importance of each principle in the optimization process. The Bayes principle seeks to minimize the average risk across all traffic participants. Let $S_R$ denote the set of all traffic participants, and let $R_i$ represent the risk associated with agent $i$. The principle is defined as: $J_B = \frac{\sum_{i=1}^{|S_R|} R_i}{|S_R|}$. This component ensures that the agent strives to reduce the overall risk in the environment, regardless of which participant bears it. To promote fairness in risk distribution, the Equality principle penalizes large disparities in individual risk. It is defined as: $J_E = \frac{\sum_{i=1}^{|S_R|} \sum_{j=i}^{|S_R|} |R_i - R_j|}{\sum_{k}^{|S_R|-1} k}$. This term discourages policies that significantly reduce total risk by disproportionately increasing the risk for specific individuals. The Maximin principle focuses on minimizing the most severe harm, regardless of its likelihood. This decouples collision probability from the severity of outcomes. Let $S_H$ represent the set of all harm values $H_i$, then the principle is defined as: $ J_M =[ \max_{H_i}(S_H)]^\gamma$, with a discount factor $\gamma  \geq 1$ that controls the sensitivity to worst-case harm. This principle distinguishes between frequent low-severity events and rare but potentially catastrophic outcomes, ensuring that decisions are sensitive to the most vulnerable participants. \par

In contrast, the \textit{selfish} mode considers only the risk to the ego vehicle. Instead of using Equation~\eqref{eq:cost}, the cost function simplifies to:

\begin{equation}
    J_{\text{ego}} = \omega_S J_S
    \label{eq:ego cost}
\end{equation}
with $J_S = \frac{\sum_{i=1}^{|S_R|} R_{i, e}}{|S_R|}$. Here, $R_{i,e}$ denotes the risk imposed on the ego vehicle by other participants. Although the form of $J_S$ resembles that of $J_B$, the two quantify different objectives. While $J_B$ accounts for the risks experienced by all traffic participants, $J_S$ ignores the effects of the ego vehicle on others and focuses solely on minimizing its own risk, effectively optimizing for self-preservation. \par

To further protect VRUs, the current cost function can be extended to incorporate preferences for lower VRU risk within its three components. In this paper, we adopt an equal treatment of all road users. However, prioritization of VRUs can be achieved by introducing weighting factors to the risk and harm terms associated with them, thereby placing greater emphasis on their protection.

\subsection{Prioritized experience replay}
\label{sec:per}

During training, we observed that risk-related cost values are typically close to zero, with only occasional spikes to high values. This sparsity in ethical cost signals poses a challenge: under standard experience replay, these rare but informative transitions are underrepresented, preventing the agent from effectively learning ethically aware behaviors. \par

To address this issue, we adopt PER, which improves learning efficiency by focusing on the most informative transitions, those with higher temporal-difference (TD) errors. PER assigns a priority to each transition $ d_i = (s_t, a_t, r_t, s_{t+1})$ based on its TD error $\delta_i$, where the priority is defined as  $ p_i \propto |\delta_i| $. Transitions are then sampled with probability:

\[
    P_i = \frac{p_i^\alpha}{\sum_j p_j^\alpha},
\]

where $\alpha \in [0,1]$ controls the degree of prioritization, with $\alpha = 0$ corresponding to uniform sampling. To correct the bias introduced by non-uniform sampling, PER applies importance sampling weights:

\[
    \omega_i = \frac{1}{{(N P_i)}^\beta},
\]
where $N$ is the buffer size and $\beta \in [0,1] $ anneals toward 1 over training to ensure asymptotically unbiased updates. To balance the influence of normal driving behavior (modeled via rewards) and ethical decision-making (modeled via costs), we compute transition priorities using a dynamic weighted combination of the reward TD error $\delta_{r, i}$ and cost TD error $\delta_{c, i}$:

\[
    p_i = \omega_r \cdot |\delta_{r, i}| + \omega_c \cdot |\delta_{c, i}|, 
\]
where the weights $\omega_r$ and $\omega_c$ are dynamically adjusted based on the relative magnitudes of the two error signals. Specifically, we define:

\[
  \omega_r = \frac{r_e}{1+r_e}, 
  \quad\quad
  \omega_c = \frac{1}{1+r_e},
\]
with the error ratio $r_e = \delta_{r,i}/\delta_{c,i}$, clipped to the range [0.2, 5] to ensure training stability. This adaptive prioritization mechanism enables the agent to focus learning on whichever signal, reward or cost, is more informative at a given time. As a result, the policy evolves with a balanced emphasis on both task performance and ethical responsibility. The overall algorithm pseudocode is shown in Algorithm \ref{pseudocode}. To provide evidence of the effectiveness of this dynamic PER, we also perform an ablation study, detailed in Section \ref{sec:results}.

\section{Experiments}

\subsection{Simulation Environment}

In this work, we use the MetaDrive simulator \citep{li2022metadrive} as the training and evaluation environment due to its ability to integrate real-world driving data, enabling the simulation of realistic driving scenarios. This feature is crucial for our study, as we aim to assess the ethical decision-making capabilities of agents in settings that closely mirror real-world conditions, thereby enhancing the relevance and reliability of our evaluation. For the real-world data, we utilize the Waymo Open Dataset \citep{ettinger2021large}, which contains recordings from six cities across the United States. It includes a wide variety of challenging scenarios, such as unprotected turns, merges, lane changes, and complex intersection interactions, making it particularly well-suited for our task. During training and evaluation, other vehicles (besides the ego vehicle) are reactive, meaning that they primarily follow the trajectories provided in the Waymo Dataset. However, when a potential rear-end collision is detected, their behavior is replaced with the intelligent driver model (IDM) \citep{treiber2000congested} and MOBIL model \citep{kesting2007general}.

\subsection{RL Setups}
\label{sec:obs and act}
The setup for safe RL training is important for a converged policy. Here, we discuss important components for safe RL training. \par

\paragraph{Observation and Action Space} In our designed scenarios, we focus on ethical decision-making and assume the existence of a reliable perception module. This module provides accurate, real-time information about surrounding traffic participants, including vehicles, pedestrians, and cyclists. As a result, our RL agent is dedicated solely to decision-making rather than perception. To support this task, we define the observation of the agent at each timestep $t$ as $o_t$. We use the Frenet coordinate system instead of Cartesian coordinates for both the observation and action spaces. The observation $o_t$ is a stacked vector composed of four components: $o_{ov, t}$, containing features of the ego vehicle; $o_{on, t}$, representing navigation-related features; $o_{sv, t}$, describing the surrounding vehicles; and $o_{sp, t}$, which accounts for other traffic participants such as pedestrians and cyclists.

\begin{equation}
    o_t = \left( (o_{ov, t})^{\top}, (o_{on, t})^\top, (o_{sv, t})^\top, (o_{sp, t})^\top \right) ^ \top.
    \label{eq:obs all}
\end{equation}

The ego vehicle observation vector $o_{ov, t}$ contains the following normalized features:

\begin{equation}
    o_{ov, t} = \left( \frac{d_{l,t}}{d_w}, \frac{d_{r,t}}{d_w}, \frac{v_{ev, t}}{v_{max}}, \frac{[\psi_{ev, t}-\chi_{ev, t}]_{-\pi}^{\pi}}{\pi},  \dot{\psi}_{ev, t} \right)^\top,
    \label{eq:obs ev}
\end{equation}
where $d_{l,t}$ and $d_{r,t}$ are the lateral distances from the ego vehicle to the left and right road boundaries, normalized by the road width $d_w$. The term $v_{ev, t}$ is the ego speed, normalized by the maximum allowable speed $v_{max}$.  The heading angle $\psi_{ev, t}$ and yaw rate $\dot{\psi}_{ev, t}$ describe vehicle orientation, while $\chi_{ev, t}$ is the heading of the road at the location of the ego vehicle. \par

Because path planning requires navigation information, we assume that future waypoints are available. To enhance robustness, we include two future waypoints. The navigation observation vector is defined as $o_{on, t} = \left( (o_{on, t}^{wp1})^\top, (o_{on, t}^{wp2})^\top \right)$, where each waypoint observation vector is given by:

\begin{equation}
    o_{on, t}^{wp} = \left(\frac{\Delta d_{d,t}}{d_{scale}}, \frac{\Delta d_{s,t}}{d_{scale}}, \frac{\kappa_{l,t}}{d_{norm}}, \frac{\psi_{l,t}}{\psi_{norm}}  \right)^\top,
\end{equation}

with $\Delta d_{d,t}$ and $\Delta d_{s,t}$ represent the lateral and longitudinal distances to the waypoint.  $\kappa_{l,t}$ and $\psi_{l,t}$ describe the curvature and heading of the lane. Constants $d_{scale}$, $d_{norm}$, and $\psi_{norm}$ normalize the values.

To model interactions with nearby traffic, we consider the eight closest surrounding vehicles. The observation is represented as $o_{sv, t} = \left( (o_{sv,t}^1)^\top, \ldots, (o_{sv,t}^8)^\top \right)$. Each surrounding vehicle $i$ is described by:

\begin{equation}
    o_{sv,t}^i = \left(\frac{\Delta d_{d,t}^{sv_i}}{d_{scale}^{sv}}, \frac{\Delta d_{s,t}^{sv_i}}{d_{scale}^{sv}}, \frac{ \Delta v_{sv_i,t}^{d}}{v_{max}}, \frac{\Delta v_{sv_i,t}^{s}}{v_{max}}  \right),
\end{equation}
where $\Delta d_{d,t}^{sv_i}$ and $\Delta d_{s,t}^{sv_i}$ are the lateral and longitudinal distances to the $i$-th vehicle. $\Delta v_{sv_i,t}^{d}$ and $\Delta v_{sv_i,t}^{s}$ are the relative velocities in lateral and longitudinal directions. The scaling factor $d_{scale}^{sv}$ reflects the perception range for nearby vehicles.

In addition to vehicles, we include pedestrians and cyclists in the observation space. The structure is the same as that used for vehicles, but we limit the number of tracked agents to four. The observation vector is defined as $o_{sp, t} = \left( (o_{sp,t}^1)^\top, \ldots, (o_{sp,t}^4)^\top \right)$. Each element follows the same format as $o_{sv,t}^i$, ensuring consistency in representation. All parameters used in the observation space are summarized in Table~\ref{tab:Hyperparameters used for the reward function setup}.\par

Since the objective of the agent is path planning, it must generate a smooth polynomial trajectory in the Frenet space at each timestep $t$. As discussed in Section \ref{sec:problem}, the action $a_t$ is defined as $a_t = \left(T_{t}, D_{d, t}, v_{s, t}\right)$. To ensure the stability of the generated trajectories, we constrained the action with $T_t \in [0, 2]$, $D_{d, t}\in [0, d_w/2]$, and $v_{s, t}\in [0, v_{max}]$. The actions are normalized during training to enhance learning stability. Given these parameters, the agent generates a polynomial trajectory using the path planner. More details are provided in Section~\ref{sec:path generate and following}.

\paragraph{Reward Function}
\label{sec: reward}

To guide the RL agent toward the desired driving behavior, a well-designed reward function is essential. In our setup, we define the reward at time step $t$ as:

\begin{equation}
    R_t = \sum _0^T R_\tau + \omega_{tr} R_{tr, t} ,
    \label{eq:reward all}
\end{equation}
with the first term representing the total reward accumulated over the planning horizon $T$, and $R_{tr, t}$ indicates the penalty term derived from the longitudinal jerk of the planned trajectories at time step $t$. It helps promote smoother driving. As shown in Table \ref{tb:reward}, this penalty sums the longitudinal jerk over all $n$ planned steps. The immediate reward $R_\tau$ at each time step $\tau$ is defined as:

%\begin{equation}
%    R_\tau = \omega_v r_{v,\tau} + \omega_\psi r_{\psi, \tau} + \omega_e r_{e, \tau} + \omega_p r_{p,\tau} + \Delta R_{\text{terminal}},
%\end{equation}

\begin{equation}
    R_\tau = \omega_v r_{v,\tau} + \omega_p r_{p,\tau} + \Delta R_{\text{terminal}},
\end{equation}

with three components, the speed reward $r_{v, \tau}$, the progress reward $r_{p,\tau}$, and a one-time terminal reward $\Delta R_{\text{terminal}}$. The weights $\omega_v$ and $\omega_p$ control the importance of the speed and progress terms. The speed reward encourages the agent to maintain a desired speed $v_{\text{des}}$. It uses a binary direction flag $\rho_{pr} \in \{ -1, +1\}$, which indicates whether the vehicle is moving in the correct direction. This helps enforce directionality and penalizes driving the wrong way. The progress reward gives the agent a small positive reward $\Delta s$ for moving forward in the correct direction. This incentivizes consistent progress along the path. The terminal reward $\Delta R_{\text{terminal}}$ is given only when the agent reaches a terminal condition: 

\begin{equation}
  \Delta R_{\text{terminal}} =
  \left\{
  \begin{array}{ll}
    R_{\text{success}}   & \quad \text{Arrival at destination}, \\[2pt]
    R_{\text{out}}       & \quad \text{Out of road}, \\[2pt]
    R_{\text{collision}} & \quad \text{Collision},
  \end{array}
  \right.
\end{equation}
where the success reward is set to $R_{\text{success}} = 25$, the off-road penalty to $R_{\text{out}} = -15$, and the collision penalty depends on the type of object involved: $R_{\text{collision}} = -10$ for collisions with other vehicles and $R_{\text{collision}} = -15$ for collisions with vulnerable road users such as pedestrians or cyclists. Table \ref{tb:reward} summarizes the reward components and their corresponding functions. The hyperparameters used in this reward formulation are listed in Table~\ref{tab:Hyperparameters used for the reward function setup}. Overall, this reward function encourages safe, efficient, and goal-oriented behavior. It balances competing objectives such as maintaining speed, staying on track, and avoiding abrupt maneuvers or collisions. The modular structure also allows easy adjustment of weights and terms for different driving scenarios and training goals.

\begin{table}
  \caption{Reward components used for RL agent.}
  \label{tb:reward}
  \centering
  \begin{tabular}{ll}
    \toprule
    Reward term     & Functions\\
    \midrule
    Longitudinal jerk penalty & $R_{tr, t} = - 0.1 n \cdot \sum |\ddot{v}_{s, \tau}| $ \\
    Speed reward & $r_{v, \tau} = \left(1 - \frac{|v_\tau - v_{\text{des}}|}{v_{\text{des}}}\right) \cdot \rho_{pr}$\\
    Progress reward     & $r_{p,\tau} = \Delta s \cdot \rho_{pr}$  \\
    \bottomrule
  \end{tabular}
\end{table}

\subsection{Low-level Control}
\label{sec:path generate and following}

To execute the high-level motion targets generated by the RL agent, we employ a dedicated execution layer for trajectory planning and following. This modular approach allows us to separate the decision-making pipeline and decouple high-level decision-making from low-level control, thereby simplifying the overall problem and accelerating convergence. In the following, we focus on the low-level control module, which consists of a path planner and a path follower.

\subsubsection{Path Planner} To ensure the comfort of the generated trajectory $\xi_{e, t}$ at time step $t$, a quartic polynomial in the longitudinal direction $l$ and a quintic polynomial in the lateral direction $d$ are employed, such that, $\xi_{e, t} = \{l_{e, \tau}, d_{e, \tau}\}$. This formulation follows established practice, as it guarantees the continuity of both longitudinal and lateral accelerations, thereby contributing to a smooth and comfortable trajectory.

\begin{equation}
  \setlength{\arraycolsep}{0pt}
  \left\{ \begin{array}{ l l }
    l_{e, \tau} &{}= a_0 + a_1 \tau + a_2 \tau^2 +a_3 \tau^3 + a_4 \tau^4,\\
    d_{e, \tau} &{}= b_0 + b_1 \tau + b_2 \tau^2 +b_3 \tau^3 + b_4 \tau^4 + b_5 \tau^5,
  \end{array} \right.
  \label{eq:traj}
\end{equation}
for $\tau \in \{0, \Delta \tau, 2\Delta \tau, ..., T\}$, where $\{ a_1,...,a_4\}$ and $\{ b_1,...,b_5\}$ denote the polynomial coefficients in the longitudinal and lateral directions, respectively. To solve equation \eqref{eq:traj}, both the initial vehicle state $[l_0, \dot{l}_0, \ddot{l}_0, d_0, \dot{d}_0, \ddot{d}_0]$ and the terminal state at the planning horizon $T$, given by, $[\dot{l}_T, \ddot{l}_T, d_T, \dot{d}_T, \ddot{d}_T]$, must be specified. \par

Following the approach in \citep{werling2010optimal}, and under the assumptions $\ddot{l}_T = 0$, $\dot{d}_T = 0$, and $\ddot{d}_T = 0$, the resulting trajectory minimizes jerk, thereby enhancing passenger comfort. In this framework, vehicle trajectories are generated to achieve a target longitudinal speed $\dot{l}_T$ and a desired lateral position $d_T$. To enable the RL agent to adaptively control driving behavior in response to dynamic traffic environments, the planning horizon $T$ is treated as an action output of the agent. This design allows the agent to flexibly determine the duration of each trajectory segment. With the solutions appended in Section \ref{sec:solutions} for equation \eqref{eq:traj}, a sequence of waypoints is produced, which are then tracked by a path-following controller. To discourage the generation of uncomfortable or impractical trajectories, a penalty term on longitudinal jerk is incorporated into the reward function, as detailed in Section \ref{sec: reward}. This encourages the agent to generate smooth and context-appropriate trajectories.

\subsubsection{Path Follower} To ensure accurate trajectory tracking, the vehicle is equipped with two controllers: a PID controller for longitudinal control and a Stanley controller \citep{thrun2006stanley} for lateral control. The longitudinal controller regulates the speed by continuously minimizing the error between the current and target longitudinal velocities. This error, denoted by $e(\tau)$, is used to compute the required acceleration through a combination of proportional ($P$), integral ($I$), and derivative ($D$) terms. The resulting control law is expressed as: 

\begin{equation}
    a(\tau) = K_p e(\tau) + K_i\int_0^\tau e(\tau) + K_d \frac{\partial}{\partial \tau} e(\tau),
\end{equation}

where $K_p$, $K_i$, and $K_d$ are the proportional, integral, and derivative gain coefficients, respectively. \par

For lateral control, the steering angle is used as the control output, and a Stanley controller is employed for this purpose. The Stanley method is a geometric path-tracking controller that gained prominence after contributing to the victory of Stanford in the 2006 DARPA Grand Challenge \citep{thrun2006stanley}. It is particularly effective because it accounts for both lateral displacement and heading errors. The steering command $\delta(\tau)$ generated by the controller is defined as: 
    
\begin{equation}
\delta(\tau) = \theta_{p}(\tau) + \tan^{-1} \left\{\frac{k_v \;d_{f}(\tau)}{v(\tau)}\right\},
\end{equation}

where $\theta_{p}(\tau)$ represents the heading error, and the second term compensates for the lateral displacement error ${d_f}$, measured from the front axle center to the nearest point on the reference trajectory. The term $k_v$ is a tuning parameter that normalizes the influence of the lateral error relative to the vehicle longitudinal speed $v_s(\tau)$, effectively defining a speed-dependent lookahead distance. \par

By combining the PID-based longitudinal controller with the Stanley-based lateral controller, the system produces a unified control command $\left\{a(\tau), \delta(\tau)\right\}$, comprising the acceleration and steering inputs at each step. This coupling ensures coordinated trajectory tracking in longitudinal and lateral dimensions, enabling smooth and accurate vehicle control.

\subsection{Baselines}

In addition to evaluating the proposed EthicAR method, we conduct ablation studies to assess the contribution of each component. The following variants are used for comparison:

\begin{itemize}
    \item \textcolor[HTML]{699DCB}{EthicAR}: The full proposed agent, which employs the LSTM-based SACLag algorithm to facilitate rare ethical decision-making and incorporates dynamic PER for policy updates.
    \item \textcolor[HTML]{6AAA81}{EthicAR w/o PER}: This variant removes dynamic PER during the policy and value network updates, allowing us to evaluate the impact of PER on performance. 
    \item \textcolor[HTML]{E68785}{SACLAG}: This version uses the standard SACLag algorithm along with the proposed dynamic PER, but excludes the LSTM structure in the network, enabling an analysis of the contribution of the LSTM.
    \item \textcolor[HTML]{E39A3C}{LSTMSAC}: In this variant, the safety constraints of the SACLag framework are removed. Instead, the cost function described in Section~\ref{sec:cost} is incorporated directly into the reward function \eqref{eq:reward all}, to assess the importance of explicitly modeling safety via constrained optimization.
\end{itemize}

As described in Section~\ref{sec:cost}, we define two cost functions, ethical and selfish, based on distinct decision-making objectives. During training, each model is trained under both modes to evaluate its behavior under different objectives. Specifically, for EthicAR, EthicAR w/o PER, and SACLag, we train two separate agents using the ethical and selfish cost functions, respectively. In contrast, for the LSTMSAC agent, the cost functions are incorporated directly into the reward function as penalty terms, allowing it to be trained as a standard RL agent without explicit safety constraints. Based on this design, we define three distinct training modes for the LSTM-SAC agent: (1) a selfish mode, (2) an ethical mode, and (3) a standard mode, where no cost function is included in the reward. The standard mode serves as a baseline for comparison, as it excludes any explicit ethical or safety consideration. This version of LSTMSAC focuses solely on conventional driving objectives, such as performance, comfort, safety, and efficiency, reflecting typical reinforcement learning behavior without ethical constraints. \par

\begin{figure}
    \centering
    \includegraphics[width=\linewidth]{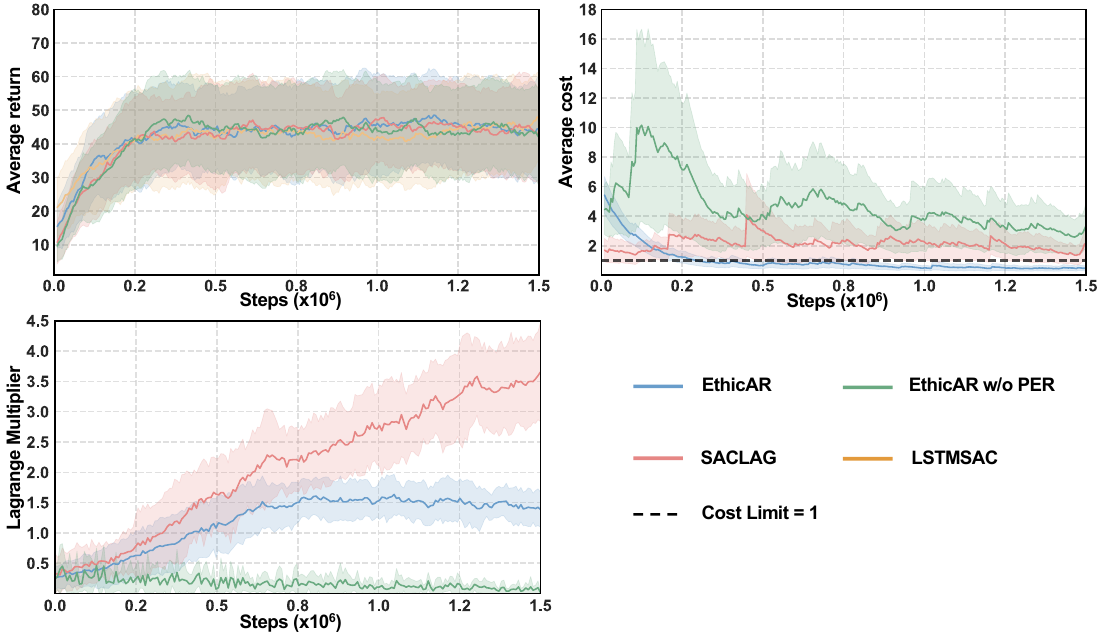}
    \caption{Training rewards and costs of different agents, together with the evolution of Lagrange multipliers in safe RL. Results are obtained with the cost mode set to \textit{ethical} and the cost limit fixed at 1.}
    \label{fig:reward_plot}
\end{figure}

Each agent was trained for 1.5 million timesteps across multiple runs using different random seeds. The simulation environment operates with a timestep interval of 0.1 seconds with 1300 real-world scenarios from the Waymo dataset. To evaluate performance under varying safety constraints, we define six different cost limits: $\eta = \{0.1, 0.3, 0.6, 0.75, 1, 2\}$, which represent varying levels of strictness. Figure~\ref{fig:reward_plot} presents the results for all agents trained under the ethical mode with the cost limit set to $\eta = 1.0$, showing both the cumulative reward and corresponding cost over time. \par

From the average reward plots in Figure~\ref{fig:reward_plot}, all agents appear to converge to a similar reward level. However, the average cost plots reveal that only the EthicAR agent successfully converges while satisfying the cost constraint. This indicates that, unlike EthicAR, the baseline agents exhibit limitations in enforcing cost-aware behavior during training. In particular, for EthicAR w/o PER, as the costs are near zero for most of the training steps, the agent has a hard time to learn effective cost-constrained behavior in the absence of dynamic PER. The SACLAG agent performs better in this regard, maintaining a relatively low cost, yet still fails to meet the cost limit consistently. \par

The evolution of Lagrange multipliers in Figure~\ref{fig:reward_plot} during training exhibits consistent trends across agents. According to Equation~\eqref{eq:safe rl}, the Lagrange multiplier $\lambda$ regulates the trade-off between reward maximization and constraint satisfaction. For the EthicAR agent, $\lambda$ initially rises as the agent explores and frequently violates cost constraints, thereby enforcing constraint adherence. As training progresses, $\lambda$ oscillates to balance rewards and costs before eventually plateauing at levels that maintain constraint satisfaction without excessive penalization. In contrast, the SACLag agent, while able to detect rare cost violations through dynamic PER, fails to converge to stable constraint levels due to the absence of the LSTM architecture. Consequently, it cannot achieve the desired balance between reward optimization and cost constraint enforcement. For EthicAR w/o PER, the scarcity of cost signals prevents the agent from learning meaningful behaviors related to cost limitation. As a result, $\lambda$ remains near zero throughout training, and the agent fails to develop sensitivity to cost constraints. These findings highlight the critical role of both the LSTM architecture and dynamic PER in enabling effective learning under explicit cost constraints.

\section{Evaluation and results}
\label{sec:results}

To further assess the performance of the proposed agent and baseline methods, we evaluate all models on 75 unseen, real-world scenarios drawn from the Waymo dataset. These scenarios were held out during training to ensure an unbiased comparison. The evaluation results are presented and analyzed in this section.

\subsection{Constraint satisfaction}

\begin{figure}
    \centering
    \includegraphics[width=\textwidth]{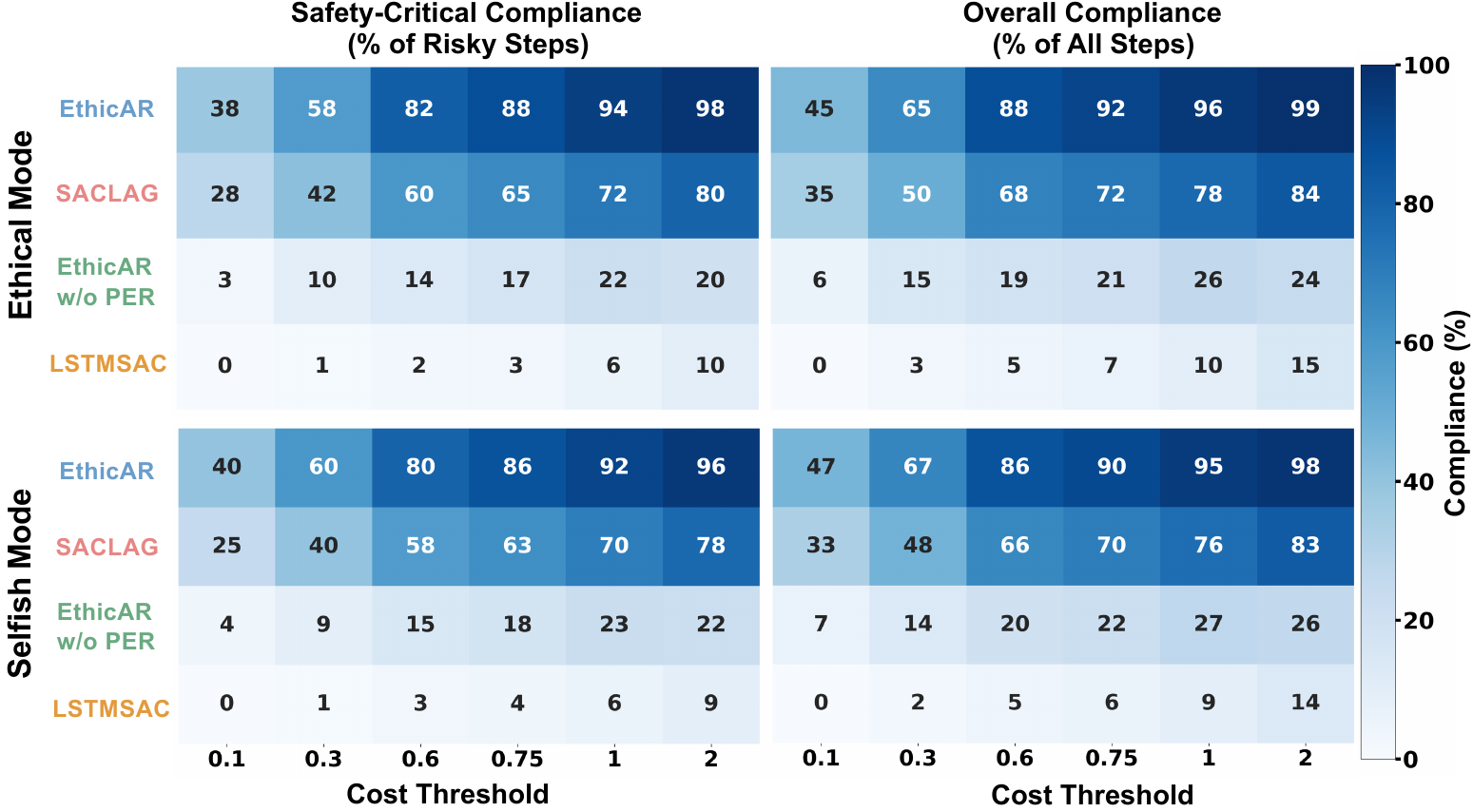}
    \caption{Constraint compliance of Safe RL agents across thresholds and modes. Compliance is shown for four agents (EthicAR, SACLag, EthicAR w/o PER, and LSTMSAC) across six cost thresholds ($\eta = 0.1\sim2$). Results are reported for all steps (right panels) and risky steps only (with a risk greater than 0), under two training modes: ethical (top) and selfish (bottom). EthicAR achieves the highest compliance, particularly in the suitable threshold range ($\eta = 0.6\sim1$), while SACLag learns partially, and EthicAR w/o PER fails to adapt to rare but critical violations. The LSTMSAC agent shows near-zero compliance across all settings. The selfish mode yields similar outcomes, confirming robustness across configurations.}
    \label{fig:compliance}
\end{figure}

First, we evaluate how well different agents satisfy cost constraints across two modes using a compliance metric, reporting both overall and risk-conditioned compliance. Compliance is defined as the percentage of steps in which the cost remains below the threshold $\eta$. Intuitively, higher compliance indicates that the agent more frequently operates within ethical safety limits. Since constraint enforcement is only meaningful under conditions where violations can occur, we measure compliance in two ways: (1) across all steps, which reflects overall stability over episodes, and (2) restricted to risky steps, where the risk value exceeds zero. We compare our EthicAR agent against three baselines, SACLAG, EthicAR w/o PER, and LSTMSAC, across six constraint thresholds in both ethical and selfish modes. Results are shown in Figure~\ref{fig:compliance}. \par

From Figure \ref{fig:compliance}, we identify $\eta \in [0.6,1]$ as the suitable operating range. Thresholds below 0.6 are unrealistically strict, such that almost no agent can comply, while thresholds above 1 are too permissive to meaningfully enforce ethical behavior. In practice, the interval $[0.6,1]$ balances feasibility and enforcement, reflecting scenarios where the agent must reliably avoid severe violations while tolerating small, unavoidable costs due to environmental uncertainty. Accordingly, we focus on cost thresholds $\eta = \{0.6, 0.75, 1 \}$ in the following analysis. \par

Across both modes, EthicAR consistently achieved the highest compliance, exceeding $88\sim96\%$ on all steps and $90\sim95\%$ on risky steps in the suitable threshold range. This indicates that EthicAR not only maintains cost constraints under normal operation but also reliably avoids rare but critical violations. SACLAG achieved moderate compliance ($68\sim78\%$ on all steps; $60\sim70\%$ on risky steps), suggesting partial success in constraint learning but limited robustness in rare-event cases. EthicAR w/o PER performed poorly, with compliance below $30\%$ on risky steps, confirming that prioritized experience replay is essential for learning ethical behavior when violations are rare. Finally, the LSTMSAC agent remained at floor levels ($<15\%$), highlighting the gap between structured Safe RL and naive behavior. A key observation is that overall compliance can appear inflated due to the predominance of trivially safe steps, whereas risky-step compliance better reflects an agent’s ability to enforce constraints in critical situations. The similarity of results between ethical and selfish modes further shows that these findings are robust across training conditions. \par

Taken together, these results show that EthicAR successfully enforces ethical constraints even under rare-event conditions, while SACLAG only partially succeeds and alternative approaches fail. The consistency across modes highlights both the generality of these findings and the importance of combining LSTM-based architectures with dynamic PER for safety-critical, ethics-aware learning.

\subsection{Risk evaluation}

\begin{figure}
    \centering
    \includegraphics[width=\textwidth]{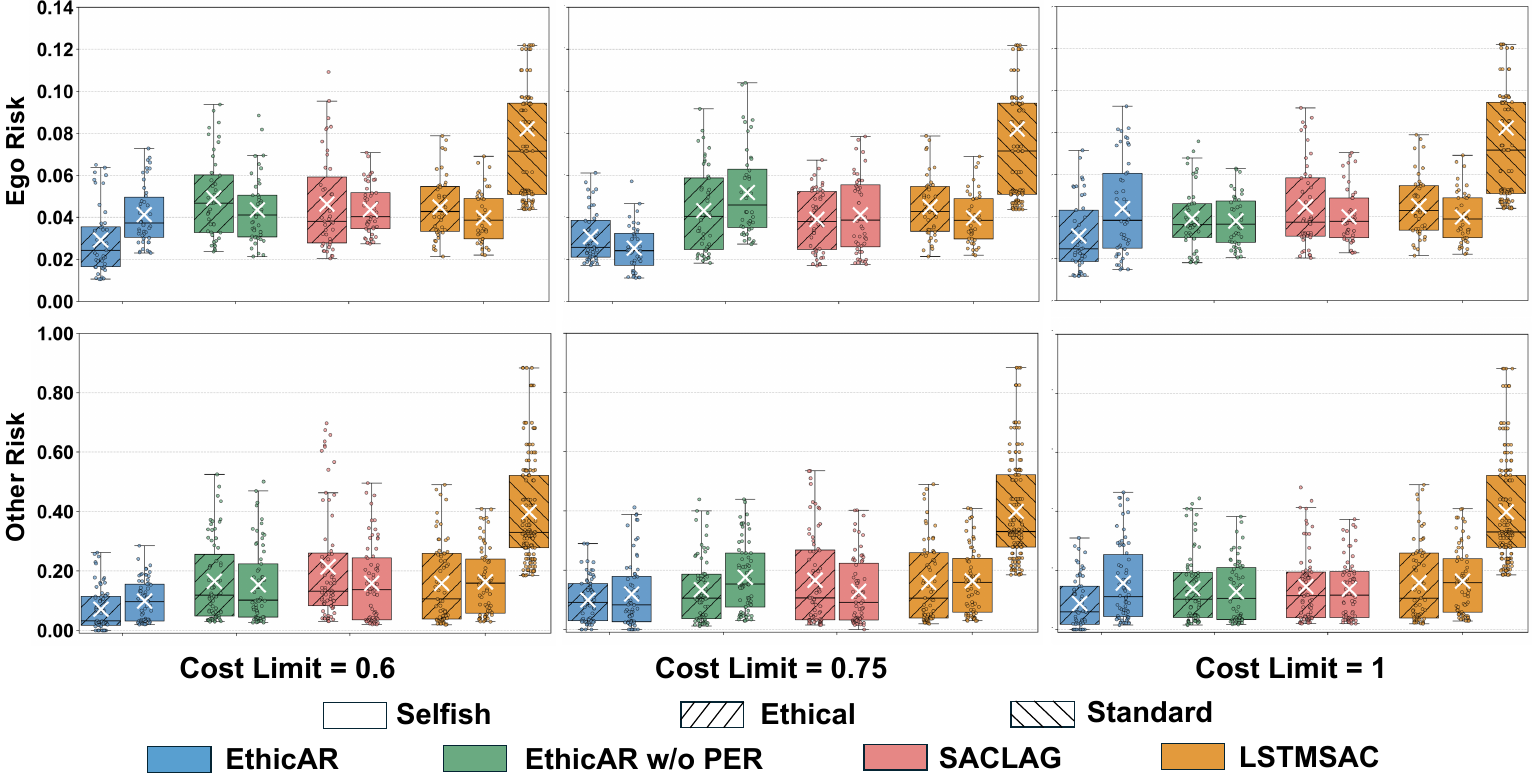}
    \caption{Risk distributions of EthicAR and other agents evaluated under three different cost limits. The first row illustrates the ego vehicle risk, while the second row depicts the risk to other traffic participants.}
    \label{fig:risk_all}
\end{figure}

Then, to further evaluate the effectiveness of the ethical considerations in EthicAR and other agents, we logged the risks discussed in Section \ref{sec:risk_cal} during evaluation. These include the risk to the ego vehicle, referred to as ego risk, and the risk to other traffic participants, referred to as other risk. The results are shown in Figure \ref{fig:risk_all} and Table \ref{tab:ego_risk} and \ref{tab:other_risk}, where each agent is evaluated in two modes across three different cost limits. An exception is the LSTMSAC agent, which does not require a cost limit for cost reduction during training, therefore, its results remain the same across all three cost limits. \par

Focusing on the other risks, results shown in the second row of Figure \ref{fig:risk_all} and Table \ref{tab:other_risk}, we observe that LSTMSAC in standard mode exhibits the highest risk to other traffic participants. This is expected, as its reward function does not incorporate any ethical considerations regarding the risk distributions among road users. In contrast, with the EthicAR agent, we observe that reducing the cost limit from 1.0 to 0.75 and then to 0.6 leads to a corresponding reduction of risks in both the selfish and ethical modes. This indicates that EthicAR, in both modes, effectively limits the costs associated with ego and other risks. When the cost limit is set to 0.6, the other risk for both modes is substantially lower than that of the other agents. Specifically, the ethical mode achieves the lowest other risk overall, while the selfish mode still maintains a relatively low level of risk. Interestingly, even though the selfish mode only explicitly considers the risk to the ego vehicle, it nonetheless lowers the risk experienced by other road users compared to the standard mode. This unintended benefit arises from interaction dynamics: by choosing smoother, safer trajectories for the ego vehicle, the selfish mode indirectly reduces hazards for surrounding agents. \par

Considering the other agents, although EthicAR w/o PER and SACLAG can reduce risks compared to the standard mode, this reduction does not consistently improve with decreasing cost limits. This suggests that the ability to learn truly ethical behavior from the cost function is limited. A similar pattern is observed with the LSTMSAC agent: while both the ethical and selfish modes show reduced risks to others, the ethical mode occasionally results in slightly higher other risk than the selfish mode. This counterintuitive outcome further indicates ineffective training and aligns with the training performance shown in Figure~\ref{fig:reward_plot}. \par

Figure \ref{fig:risk_all} and Table \ref{tab:ego_risk} summarize the ego risk outcomes. Although these risk values remain low and vary little between agents, the differences are still informative. We designed the selfish mode to minimize risk exclusively for the ego vehicle. Surprisingly, the ethical mode EthicAR achieves similar or even lower ego risk. In particular, when the cost limits are set to 0.75 and 0.60, EthicAR yields the lowest ego risk. This finding highlights a compelling insight: by accounting for the safety of other road users, we not only lower their risk but also reduce the ego risk. In other words, cooperative traffic behavior not only benefits others but also enhances the safety of the ego vehicle.

\begin{table}
  \centering
  \caption{Ego risk distributions of EthicAR and other agents evaluated under three different cost limits.}
  \begin{tabular}{llcccc}
    \toprule
    \multicolumn{2}{c}{\multirow{2}{*}{\textbf{Agents}}} & \multicolumn{3}{c}{\makecell[c]{\textbf{Cost Limit}}}  \\
    \cline{3-5} 
    \addlinespace
     & & \textbf{0.6} & \textbf{0.75} & \textbf{1}\\
    \midrule
    \multicolumn{1}{c}{\multirow{3}{*}{\textbf{LSTMSAC}}} & Selfish & $0.040 \pm 0.012$ & $0.040 \pm 0.012$ & $0.040 \pm 0.012$ \\
                             & Ethical    & $0.045 \pm 0.015$ & $0.045 \pm 0.015$ & $0.045 \pm 0.015$ \\
                             & Standard   & $0.082 \pm 0.054$ & $0.082 \pm 0.054$ & $0.082 \pm 0.054$\\
    \cline{2-5}
    \addlinespace
    \multicolumn{1}{c}{\multirow{2}{*}{\textbf{SACLAG}}} & Selfish & $0.044 \pm 0.012$ & $0.040 \pm 0.013$ & $0.041 \pm 0.018$\\
                             & Ethical    & $0.046 \pm 0.023$ & $0.045 \pm 0.020$ & $0.039 \pm 0.015$\\
    \cline{2-5} 
    \addlinespace
    \multicolumn{1}{c}{\multirow{2}{*}{\textbf{EthicAR w/o PER}}} & Selfish &  $0.044 \pm 0.016$ & $0.038 \pm 0.012$ & $0.052 \pm 0.021$ \\
                             & Ethical    & $0.049 \pm 0.020$ & $0.039 \pm 0.015$ & $0.043 \pm 0.020$ \\
    \cline{2-5}
    \addlinespace
    \multicolumn{1}{c}{\multirow{2}{*}{\textbf{EthicAR}}} & Selfish & $0.041 \pm 0.014 $ & $0.044 \pm 0.023 $ & $\pmb{0.025} \pm 0.011 $\\
                             & Ethical    & $\pmb{0.029} \pm 0.016$ & $\pmb{0.031} \pm 0.017$ & ${0.031 \pm 0.012}$\\
                           
    \bottomrule
  \end{tabular}
  \label{tab:ego_risk}
\end{table}

\begin{table}
  \centering
  \caption{Other risk distributions of EthicAR and other agents evaluated under three different cost limits.}
  \begin{tabular}{llcccc}
    \toprule
    \multicolumn{2}{c}{\multirow{2}{*}{\textbf{Agents}}} & \multicolumn{3}{c}{\makecell[c]{\textbf{Cost Limit}}}  \\
    \cline{3-5} 
    \addlinespace
     & & \textbf{0.6} & \textbf{0.75} & \textbf{1}\\
    \midrule
    \multicolumn{1}{c}{\multirow{3}{*}{\textbf{LSTMSAC}}} & Selfish & $0.164 \pm 0.112$ & $0.164 \pm 0.112$ & $0.164 \pm 0.112$ \\
                             & Ethical    & $0.158 \pm 0.135$ & $0.158 \pm 0.135$ & $0.158 \pm 0.135$ \\
                             & Standard   & $0.398 \pm 0.177$ & $0.398 \pm 0.177$ & $0.398 \pm 0.177$\\
    \cline{2-5}
    \addlinespace
    \multicolumn{1}{c}{\multirow{2}{*}{\textbf{SACLAG}}} & Selfish & $0.158 \pm 0.131$ & $0.137 \pm 0.102$ & $0.129 \pm 0.113$\\
                             & Ethical    & $0.213 \pm 0.193$ & $0.145 \pm 0.121$ & $0.166 \pm 0.154$\\
    \cline{2-5} 
    \addlinespace
    \multicolumn{1}{c}{\multirow{2}{*}{\textbf{EthicAR w/o PER}}} & Selfish &  $0.154 \pm 0.128$ & $0.128 \pm 0.100$ & $0.176 \pm 0.116$ \\
                             & Ethical    & $0.165 \pm 0.135$ & $0.139 \pm 0.121$ & $0.134 \pm 0.112$ \\
    \cline{2-5}
    \addlinespace
    \multicolumn{1}{c}{\multirow{2}{*}{\textbf{EthicAR}}} & Selfish & $0.099 \pm 0.070 $ & $0.157 \pm 0.131 $ & ${0.119} \pm 0.116 $\\
                             & Ethical    & $\pmb{0.071} \pm 0.074 $ & $\pmb{0.088} \pm 0.088$ & $\pmb{0.099} \pm 0.076$\\
                           
    \bottomrule
  \end{tabular}
  \label{tab:other_risk}
\end{table}

\begin{figure}
    \centering
    \includegraphics[width=\linewidth]{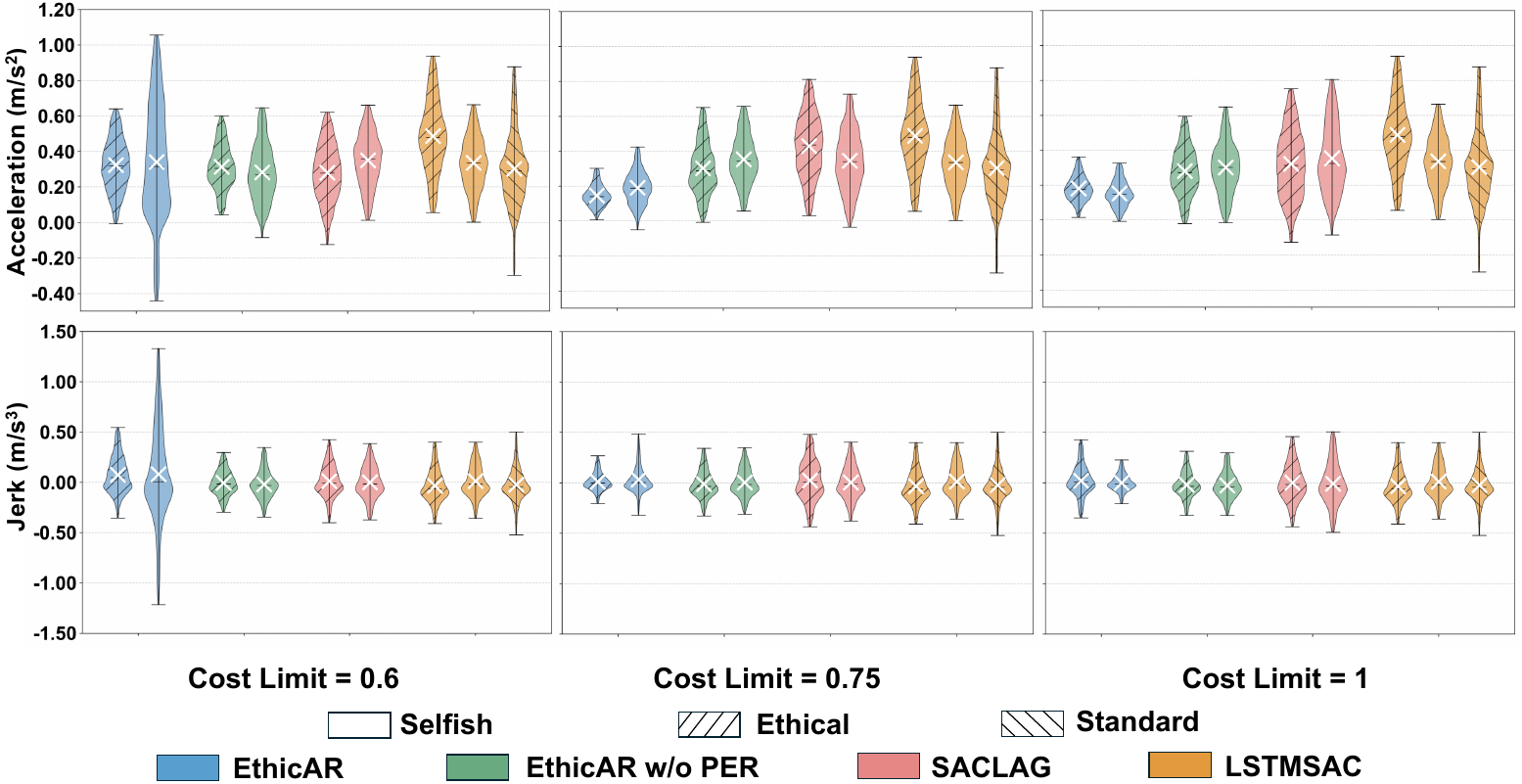}
    \caption{Distributions of acceleration (top) and jerk (bottom) for EthicAR versus baseline agents under three cost limits.}
    \label{fig:acc_plots}
\end{figure}

Apart from the risk analysis, we also examine the dynamics of the AV to evaluate the comfort level of the proposed framework. Specifically, we analyze the accelerations and jerks of EthicAR operating in both ethical and selfish modes, as well as other agents. The results are shown in Figure \ref{fig:acc_plots}. For acceleration, values below $1\mathrm{m/s^2}$ are generally considered comfortable, while values up to $2\mathrm{m/s^2}$ are tolerable in more dynamic scenarios. In terms of jerk, comfort thresholds are typically below $1\mathrm{m/s^3}$, with values between $1$ and $3\mathrm{m/s^3}$ often associated with reduced comfort or instability \citep{de2023standards}.\par

According to Figure \ref{fig:acc_plots}, EthicAR in selfish mode shows slightly higher accelerations but still remains within the comfort zone. All other agents, across both modes, maintain accelerations within comfortable limits. When evaluating jerk, the selfish mode of EthicAR again shows some mildly discomforting actions, though these remain within the acceptable range. These results reflect the influence of the reward function (Section \ref{sec: reward}) and the trajectory planner (Section \ref{sec:path generate and following}), which together help ensure that generated trajectories and their execution stay within comfort limits.

\subsection{Worst-case analyses}

\begin{figure}
    \centering
    \includegraphics[width=\textwidth]{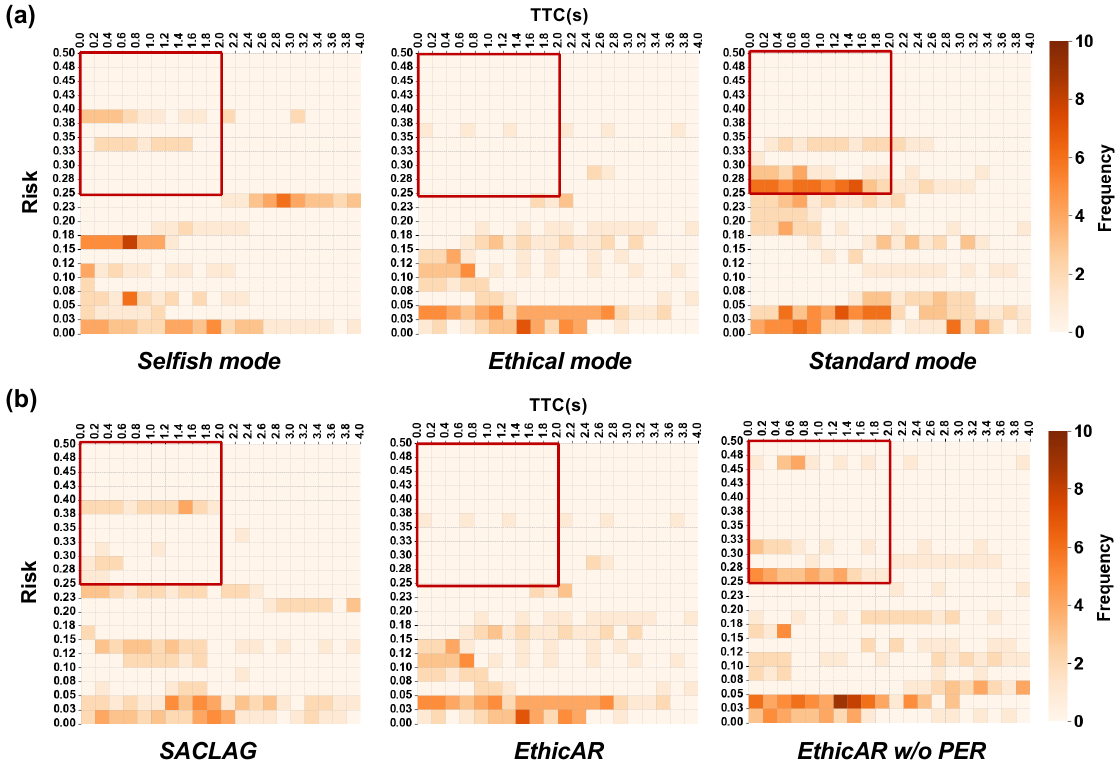}
    \caption{The heatmap of the worst-case pairings of minimum TTC and maximum other risk during the evaluation for different agents. (a) presents results for the selfish and ethical modes of EthicAR, with the standard mode included for comparison. (b) compares the ethical modes of three agents: EthicAR, SACLAG, and EthicAR w/o PER. To highlight undesirable behaviors, we mark cases with TTC less than 2 seconds and other risks greater than 0.25 using a red square in each heatmap.}
    \label{fig:heatmap_ours}
\end{figure}

After evaluating the general performance of the different agents, we proceed to assess their behavior in worst-case scenarios. For this purpose, we use the Time-to-Collision (TTC) metric, which estimates the time remaining before a collision would occur between traffic participants if they continue at their current speeds and along their current paths. TTC is a critical indicator for assessing safety. To capture these scenarios, we log the minimum TTC between the ego vehicle and other vehicles, together with the corresponding maximum other risk values at each timestep during evaluation. The underlying rationale is that a predictive and cautious agent should be able to reduce risk in advance, leading to fewer instances where both TTC is low and other risk is high. Conversely, a higher frequency of such cases indicates a failure to anticipate and mitigate danger. The results are visualized in Figure \ref{fig:heatmap_ours}, using a cost limit of 0.6 for all agents. The heatmap encodes the frequency of TTC–risk pairs: darker cells correspond to higher frequencies. For clarity, we define undesirable behavior as cases where TTC is less than 2 seconds and other risk exceeds 0.25. These critical cases are highlighted with red squares. \par

In Figure \ref{fig:heatmap_ours} (a), we compare the EthicAR agent in both ethical and selfish modes against the standard mode baseline. Within the critical area, the heatmap shows that the standard mode has the most darker cells, indicating the largest number of high-risk, low-TTC cases. This highlights its inability to mitigate danger when trained solely on the reward function \eqref{eq:reward all}. In contrast, both the ethical and selfish modes of EthicAR show markedly fewer critical cases, with the ethical mode performing best: only three isolated instances fall into the red squares. This suggests that incorporating ethical considerations, even in the selfish mode, leads to safer behavior compared to the standard mode. Figure \ref{fig:heatmap_ours}(b) extends the comparison across agents, focusing on the ethical mode of SACLAG, EthicAR, and EthicAR w/o PER. All three agents reduce the frequency of undesirable behaviors relative to the standard mode. Among them, EthicAR achieves the strongest performance, further highlighting the effectiveness of its design.\par

Using the standard mode as a baseline, we observe that EthicAR is capable of reducing risk by implicitly predicting the behavior of other vehicles—an ability embedded in the cost function described in Section \ref{sec:cost}. This cost function estimates collision probabilities by using a prediction model to forecast the trajectories of surrounding vehicles. While the agent does not receive these predicted trajectories as input, by leveraging its LSTM architecture with the current states of nearby vehicles provided as input, the agent implicitly learns to model these predictions through cost-based training. This enables EthicAR to exhibit predictive behavior, resulting in higher TTC values and ultimately lower risk during interactions with other traffic participants.

\subsection{Scenario analyses}

To provide a detailed comparison of driving behaviors, we selected four real‑world scenarios that each pose an ethical dilemma in everyday traffic. We evaluated EthicAR alongside three baseline agents under a uniform cost limit of 0.6. Each simulation advanced in 0.1s timesteps, and for every agent, we identified five critical instants at which we recorded the TTC with the most relevant road user as well as the peak other risk experienced by the ego vehicle. We then plotted TTC and risk side by side to enable clear, quantitative comparisons. In these plots, any TTC below 0s, indicating the two objects are diverging, or above 10s, indicating negligible imminent danger, is represented as inf, marking the absence of meaningful collision risk at that moment. In the following sections, we show two scenarios in turn, highlighting how EthicAR and the baselines navigate these ethical trade‑offs. Two more scenarios are appended in Section \ref{sec:more_scenario}.

\paragraph{Drive along with a cyclist on a two-way road}
\label{sec: scenario 1}

\begin{figure}
    \centering
    \includegraphics[width=\textwidth]{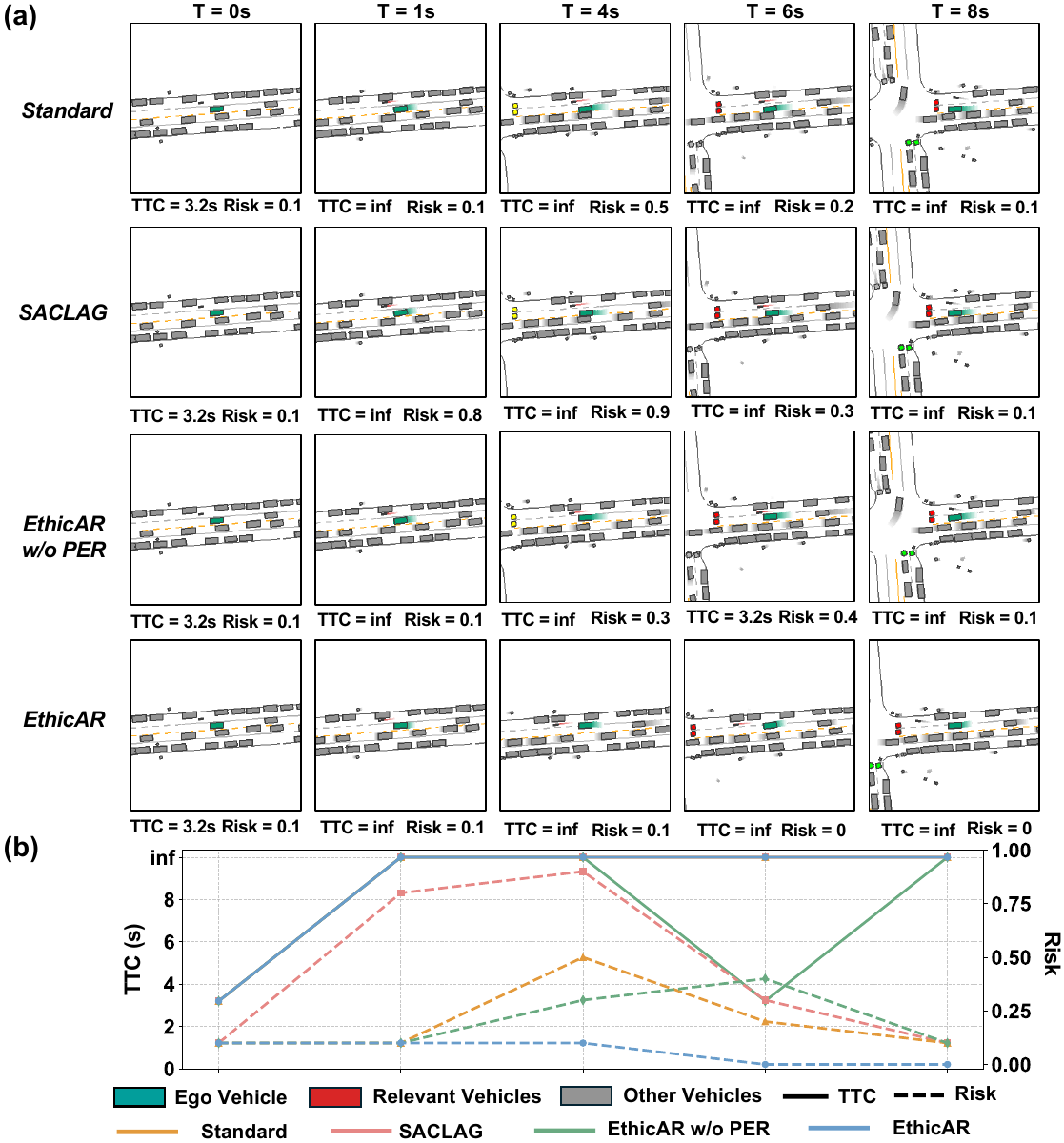}
    \caption{Real‑world driving scenario in which the ego vehicle (green) follows a cyclist (red) along a two‑way road with oncoming traffic (gray). (a) Per‑timestep behavior of EthicAR and baseline agents, showing real‑time TTC with the cyclist and the maximum other risk of the ego vehicle. (b) TTC versus maximum risk at the five selected instants, plotted for direct quantitative comparison.}
    \label{fig:ttc_risk_ep2}
\end{figure}

In the first scenario, our ego vehicle follows a cyclist on a two‑lane, bidirectional road while encountering dense traffic congestion in the opposing lane. Figure \ref{fig:ttc_risk_ep2} depicts these results: (a) shows real‑time TTC to the cyclist of EthicAR and the baseline agents alongside their peak other risk at each timestep, and (b) presents the corresponding TTC and maximum risk values at the five selected timesteps. \par

The standard‑mode agent, trained solely on efficiency and safety objectives with the reward function \eqref{eq:reward all}, exhibits its limitations in this scenario. 
Starting from a TTC to the cyclist of 3.2s and a low-risk situation, the standard mode, like other baselines, adjusts its heading to achieve a TTC of \textit{inf} by laterally diverging from the cyclist. However, with no ethical constraints, it immediately attempts to overtake once the TTC reaches \textit{inf}, closing to an uncomfortably small lateral gap. In reality, forcing an overtake on a narrow, two‑lane road with oncoming congestion violates both ethical norms and common etiquette \citep{luetge2017german}, and risks a collision either with the cyclist or opposing traffic due to the unpredictability of the cyclist. Such unsafe aggressiveness cannot be mitigated by other baselines: both the PER‑ablated EthicAR and the SACLAG agent display similarly risky overtaking attempts here. By contrast, EthicAR maintains a respectful distance from the cyclist throughout and refrains from overtaking, keeping its maximum other risk at a low 0.1 until the conclusion of the scenario, showing more socially acceptable, ethically aligned behavior suitable for everyday driving. Importantly, despite the cost function not explicitly prioritizing the minimization of risks to VRUs, EthicAR nonetheless exhibited responsible decision-making toward them. This outcome underscores the effectiveness of the proposed framework and further suggests that the protection of VRUs could be enhanced through straightforward adjustments to weight assignments.

\paragraph{Unprotected left-turn across path}

\begin{figure}
    \centering
    \includegraphics[width=\textwidth]{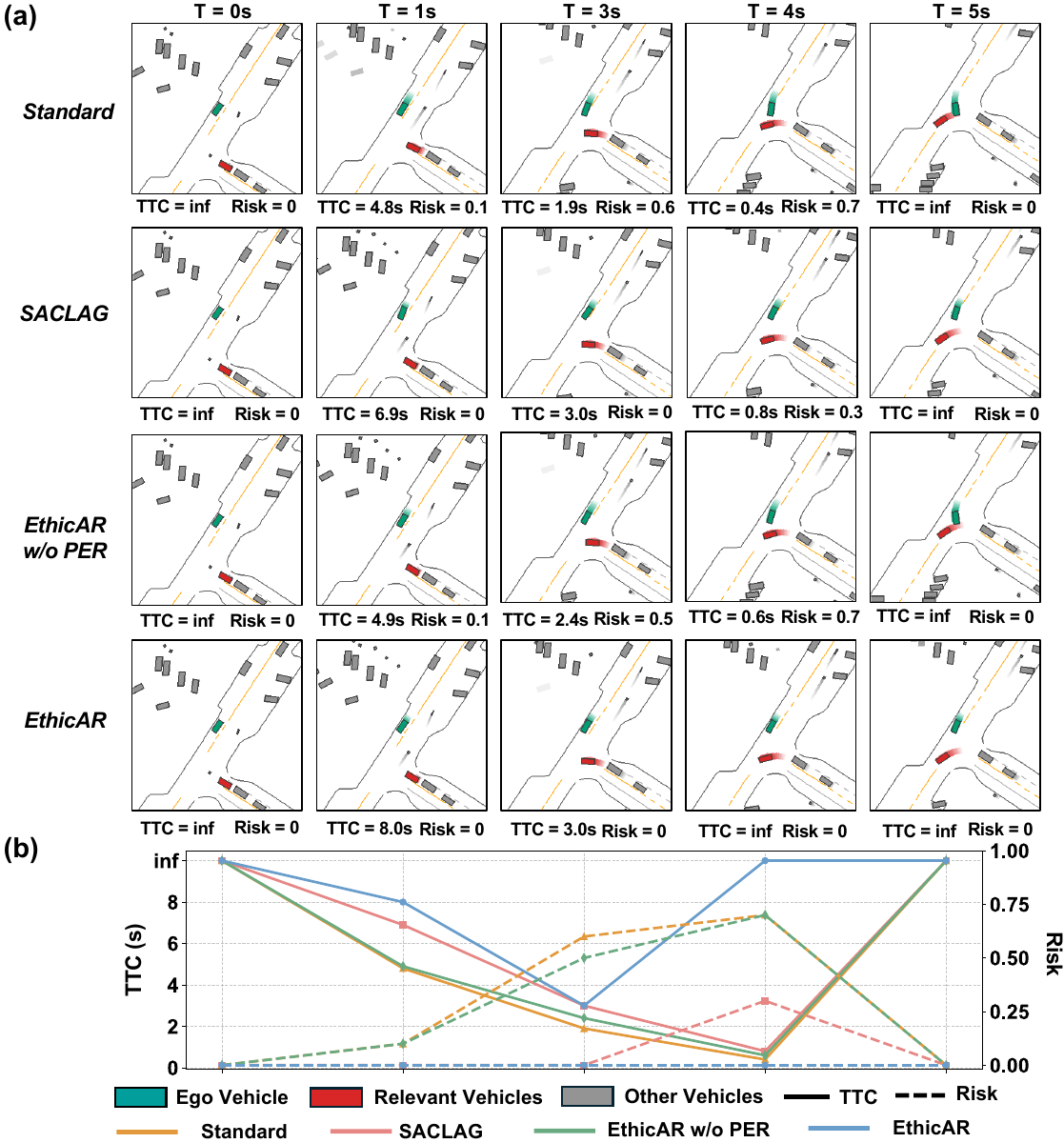}
    \caption{Real‑world driving scenario in which the ego vehicle (green) and an opposing left‑turning vehicle (red) both prepare to turn across each other, with other traffic shown in gray. (a) Per‑timestep behavior of EthicAR and baseline agents, showing real‑time TTC with the cyclist and the maximum other risk of the ego vehicle. (b) TTC versus maximum risk at the five selected instants, plotted for direct quantitative comparison.}
    \label{fig:ttc_risk_ep1}
\end{figure}

The second scenario is another very common researched social dilemma, where a vehicle turns left across traffic moving in the opposite direction without a protected signal. As shown in Figure \ref{fig:ttc_risk_ep1}, the ego vehicle, marked in green, prepares to make a left turn, while another vehicle, marked in red, also tries to make a left turn. \par

In everyday driving etiquette at an unsignalized intersection, especially when the ego vehicle approaches from the right, it technically has priority to proceed first. However, safety considerations lead all agents to \textit{active yield}. While this reflects legal norms, not all agents minimize risk effectively. For instance, for the standard mode, starting from \textit{inf} TTC, and risk of 0, and then allows the other vehicle to turn. Yet its approach remains aggressive: by $T=4\mathrm{s}$, TTC falls to just 0.4s and risk spikes to 0.7, even though no collision occurs. Such behavior is risky and unsuitable for AV deployment. EthicAR w/o PER agent displays a similar pattern. The SACLAG agent shows improved restraint: though ego speed remains relatively high at $T=4\mathrm{s}$, TTC and risk increase to about 0.3, still yielding more courteously. By contrast, EthicAR maintains the lowest risk level throughout the scenario, embodying responsible and socially considerate driving.

\section{Discussion}

In the coming decades, AVs are expected to engage in frequent and complex interactions with human drivers. Public concern remains high, particularly regarding the moral and ethical dilemmas AVs may encounter in high-stakes, split-second decision-making scenarios. As a result, the development of AVs must go beyond achieving safety and comfort, responsible and ethical decision-making is becoming an increasingly essential objective. \par

In this work, we introduce EthicAR, a hierarchical autonomous driving framework designed to minimize the overall risk to all road users and make ethically responsible decisions. EthicAR not only reacts ethically in high-risk situations but also incorporates ethical considerations into routine driving behavior, thereby proactively avoiding dangerous scenarios. The framework consists of two levels: a decision level and an execution level. The decision level is powered by an LSTM-based SACLag algorithm, enhanced with dynamic PER and trained under ethical cost constraints. The execution level ensures that planned actions are translated into smooth and comfortable vehicle control. To evaluate the contributions of each component, we conduct an ablation study examining the effects of PER, the LSTM architecture, and different cost function designs. The agents are trained using a real-world dataset in simulation, providing realistic scenarios that improve generalization and reliability. For performance evaluation, we test the trained agents on 75 previously unseen real-world scenarios. We analyze the resulting risks for both the ego vehicle and surrounding road users. Our results show that the ethical mode of EthicAR outperforms the other variants, significantly reducing overall risk for all road users compared to versions without cost constraints. Interestingly, although EthicAR is explicitly optimized to reduce risk to all road users, it also achieves lower ego risk than the selfish mode, which considers only the safety of the ego vehicle. This suggests that ethical behavior, inherently designed to protect others, also benefits the ego vehicle, due to the interactive nature of traffic dynamics. Additionally, we analyze four real-world scenarios involving ethical dilemmas. These case studies confirm that EthicAR consistently makes responsible decisions, not only toward other vehicles but also toward VRUs such as cyclists. If stronger prioritization of VRUs is required, this can be readily achieved through simple weight adjustments in the cost function. Looking ahead, extending the framework to incorporate trajectory predictions for pedestrians and cyclists, in addition to vehicles, will enable even more realistic behavior modeling and improve readiness for real-world deployment.

\section*{Acknowledgment}

 This work was supported in part by the Federal
Ministry of Research, Technology and Space of Germany and in part by the Sächsische Staatsministerium für Wissenschaft, Kultur und Tourismus in the programme Center of Excellence for AI-research Center for Scalable
Data Analytics and Artificial Intelligence Dresden/Leipzig, project ScaDS.AI.

\section*{Code availability}
All code is available via GitHub at https://github.com/DailyL/EDRL. The training of the RL agent was implemented in Python using the deep learning framework PyTorch. The code supporting the plot within this paper is also available. We have provided all the source code under the MIT license, making it open and accessible to anyone interested in building upon our work.

\bibliographystyle{unsrtnat}
\bibliography{references}

%%%%%%%%%%%%%%%%%%%%%%%%%%%%%%%%%%%%%%%%%%%%%%%%%%%%%%%%%%%%

\appendix

\section{Supplementary Material}
\label{sec:append}

\subsection{LSTM-based SACLag Pseudocode}

The pseudocode of LSTM-based SACLag with dynamic PER is presented in Algorithm \ref{pseudocode}.

\begin{algorithm}
\caption{LSTM-based SACLag with Dynamic PER}
\label{pseudocode}
\begin{algorithmic}[1]
\State Set hyperparameters: learning rates $\beta_\theta, \beta_\phi, \beta_\lambda, \beta_\alpha$, discount $\gamma$, PER exponents $\eta,\beta$, ratio bounds $[r_{\min},r_{\max}]$, small constant $\epsilon>0$, soft-update rate $\tau$, and cost threshold $d$.
\State Initialize policy $\pi_\theta$, reward critics $Q_{\phi_1}, Q_{\phi_2}$, cost critics $Q_{\psi_1}, Q_{\psi_2}$, target networks $\bar Q_{\phi_1}, \bar Q_{\phi_2}, \bar Q_{\psi_1}, \bar Q_{\psi_2}$, Lagrange multiplier $\lambda \ge 0$, temperature $\alpha$, PER buffer $\mathcal{D}$.

\For{each iteration}
    \State Sample initial state $s_0$ and initial LSTM hidden states $h_0^\pi, h_0^Q$
    \For{each environment step}
        \State $a_t \sim \pi_\theta(\cdot|s_t,h_t^\pi)$, \quad $s_{t+1} \sim P(\cdot|s_t,a_t)$
        \State  $\mathcal{D}$ $\gets$ $\mathcal{D}$ $\cup$  $\{s_t,h_t^\pi,h_t^Q,a_t,r_t,c_t,s_{t+1}\}$ with initial priority
    \EndFor
    \For{each gradient step}
        \State Sample batch $\mathcal{B} = \{(s,h^\pi,h^Q,a,r,c,s')\}$ from $\mathcal{D}$ with PER probabilities $P_i \propto p_i^\eta$
        \State Sample batch of next actions $a' \sim \pi_\theta(\cdot|s',h^{\pi\prime})$
        \State Compute TD errors:
        \[
            \delta_r = r + \gamma \Big(\min_{j=1,2} \bar Q_{\phi_j}(s',h^{Q\prime},a') - \alpha \log \pi_\theta(a'|s',h^{\pi\prime})\Big) - Q_{\phi_1}(s,h^Q,a)
        \]
        \[
            \delta_c = c + \gamma \min_{j=1,2} \bar Q_{\psi_j}(s',h^{Q\prime},a') - Q_{\psi_1}(s,h^Q,a)
        \]
        \State Compute dynamic PER weights:
        \[
          r_e = \text{clip}\!\left(\tfrac{|\delta_r|}{|\delta_c|+\epsilon}, r_{\min}, r_{\max}\right),\;
          \omega_r=\tfrac{r_e}{1+r_e},\;\omega_c=\tfrac{1}{1+r_e}
        \]
        \[
            p_i = \omega_r |\delta_r| + \omega_c |\delta_c|, \quad
            \omega_i = \frac{1}{(N P_i)^\beta}
        \]

        \State Update critics with weight $\omega_i$
        \State Update actor with Lagrangian:
        \[
          \theta \gets \theta - \beta_\theta \nabla \Big(w_i\,[\,\alpha \log \pi_\theta(a|s,h^\pi) - Q_{\phi_1}(s,a) + \lambda Q_{\psi_1}(s,a)\,]\Big)
        \]
        \State Update Lagrange multiplier:
        \[
            \lambda \gets \max\{0,\,\lambda+\beta_\lambda(Q_{\psi_1}(s,h^Q,a)-d)\}
        \]
        \State Update temperature $\alpha$ and target networks with $\tau$
        \State Update PER priorities $p$ in $\mathcal{D}$
    \EndFor
\EndFor
\end{algorithmic}
\end{algorithm}

\subsection{Hyperparameter}
\label{sec:parameter}

\begin{table}[htbp]
    \centering
    \begin{threeparttable}
    \caption{Hyperparameters used for the cost function.}
    \label{tab:Hyperparameters used for the cost function}
    \begin{tabular}{cll}
    \Xhline{1pt}
    Symbol  & Description   & Value \\
    \hline
    $c_0$  & Empirical coefficient & 4.457 \\
    $c_1$  & Empirical coefficient & 0.177\\
    $c_a$\tnote{1}  & Empirical coefficient & 0.244 or -0.431 \\
    $\omega_B$  & Bayes principle weight & 3.33\\
    $\omega_E$  & Equality principle weight & 3.33\\
    $\omega_M$  & Maximin principle weight & 3.33\\
    $\omega_S$  & Selfish mode weight & 10\\
    \Xhline{1pt}
    \end{tabular}
    \begin{tablenotes}
      \scriptsize
      \item[1] The value depends on the collision angle: if the collision occurs on the side, it is 0.244; if it occurs at the rear, it is -0.431.
    \end{tablenotes}
    \end{threeparttable}
\end{table}

\begin{comment}
\begin{table}
    \centering
    \begin{threeparttable}
    \caption{Hyperparameters used for the input states and reward function setup.}
    \label{tab:Hyperparameters used for the reward function setup}
    \begin{tabular}{cll}
    \Xhline{1pt}
    Symbol  & Description   & Value \\
    \hline
    $d_w$  & Road width & $4.5\mathrm{m}$ \\
    $v_{max}$  & Maximal allowed speed & $22.22 \mathrm{m/s}$\\
    $d_{scale}$ & Normalization constant & $50\mathrm{m}$ \\
    $d_{norm}$  & Normalization constant for curvature & 60\\
    $\psi_{norm}$  & Normalization constant for heading & 60\\
    $d_{scale}^{sv}$ & Perceive distance of vehicles & $50\mathrm{m}$ \\
    $\omega_v$  & Speed reward component weight & 0.7\\
    $\omega_p$  & Progress reward component weight & 0.1\\
    $\omega_{tr}$  & Trajectory Jerk penalty weight & 0.05\\
    $R_{\text{success}}$ & Success  reward & 25\\
    $R_{\text{out}}$ & Out of road penalty & -15\\
    $R_{\text{collision}}$\tnote{1} & Collision penalty & -10 or -15\\
    \Xhline{1pt}
    \end{tabular}
    \begin{tablenotes}
      \scriptsize
      \item[1] A collision with another vehicle incurs a penalty of -10, while a collision with a pedestrian or cyclist results in a penalty of -15.
    \end{tablenotes}
    \end{threeparttable}
\end{table}
\end{comment}

\begin{table}[htbp]
    \centering
    \begin{threeparttable}
    \caption{Hyperparameters used for the input states and reward function setup.}
    \label{tab:Hyperparameters used for the reward function setup}
    \begin{tabular}{cll}
    \Xhline{1pt}
    Symbol  & Description   & Value \\
    \hline
    $d_w$  & Road width & $4.5\mathrm{m}$ \\
    $v_{max}$  & Maximal allowed speed & $22.22 \mathrm{m/s}$\\
    $d_{scale}$ & Normalization constant & $50\mathrm{m}$ \\
    $d_{norm}$  & Normalization constant for curvature & 60\\
    $\psi_{norm}$  & Normalization constant for heading & 60\\
    $d_{scale}^{sv}$ & Perceive distance of vehicles & $50\mathrm{m}$ \\
    $\omega_v$  & Speed reward component weight & 0.7\\
    $\omega_p$  & Progress reward component weight & 0.1\\
    $\omega_{tr}$  & Trajectory Jerk penalty weight & 0.05\\
    \Xhline{1pt}
    \end{tabular}
    \end{threeparttable}
\end{table}

\subsection{Solutions for the longitudinal and lateral polynomial equations}
\label{sec:solutions}

As discussed in Section \ref{sec:path generate and following}, for the trajectory $\xi_{e, t}$ at time step $t$, we employ a quartic polynomial in the longitudinal direction and a quintic polynomial in the lateral direction. To solve these polynomials, we use the known initial vehicle state 
\[
[l_0, \dot{l}_0, \ddot{l}_0, d_0, \dot{d}_0, \ddot{d}_0]
\] 
and the terminal state at the planning horizon $T$, given by 
\[
[\dot{l}_T, \ddot{l}_T, d_T, \dot{d}_T, \ddot{d}_T].
\] 
The outputs from the RL agent provide $\dot{l}_T$, $d_T$, and $T$, while we assume $\ddot{l}_T = 0$, $\dot{d}_T = 0$, and $\ddot{d}_T = 0$. Under these assumptions, the resulting trajectory minimizes jerk, thereby enhancing passenger comfort \citep{werling2010optimal}. Therefore, the polynomial coefficients can be solved as follows.

\subsubsection*{Longitudinal Quartic Polynomial}
The longitudinal trajectory is defined as
\[
l_e(\tau) = a_0 + a_1 \tau + a_2 \tau^2 + a_3 \tau^3 + a_4 \tau^4.
\]
Using the initial state, the first three coefficients are determined:
\[
a_0 = l_0, \quad a_1 = \dot{l}_0, \quad a_2 = \frac{1}{2} \ddot{l}_0.
\]
Substituting $\tau = T$ into the polynomial and its derivatives gives the terminal constraints:
\begin{align*}
\dot{l}_T &= \dot{l}_0 + \ddot{l}_0 T + 3 a_3 T^2 + 4 a_4 T^3, \\
\ddot{l}_T &= \ddot{l}_0 + 6 a_3 T + 12 a_4 T^2.
\end{align*}
These can be expressed in matrix form to solve for the unknown coefficients $a_3$ and $a_4$:
\[
\begin{bmatrix}
a_3 \\[2pt] a_4
\end{bmatrix}
=
\begin{bmatrix}
3 T^2 & 4 T^3 \\[1pt]
6 T & 12 T^2
\end{bmatrix}^{-1}
\begin{bmatrix}
\dot{l}_T - \dot{l}_0 - \ddot{l}_0 T \\[1pt]
\ddot{l}_T - \ddot{l}_0
\end{bmatrix}.
\]
This yields $[a_3, a_4]$, completing the quartic longitudinal trajectory.

\begin{comment}
\[
a_3 = \frac{\dot{l}_T - \dot{l}_0 - \ddot{l}_0 T}{T^2} - \frac{\ddot{l}_T - \ddot{l}_0}{3 T}, 
\quad
a_4 = -\frac{\dot{l}_T - \dot{l}_0 - \ddot{l}_0 T}{2 T^3} + \frac{\ddot{l}_T - \ddot{l}_0}{4 T^2}.
\]
\end{comment}
\subsubsection*{Lateral Quintic Polynomial}
The lateral trajectory is defined as
\[
d_e(\tau) = b_0 + b_1 \tau + b_2 \tau^2 + b_3 \tau^3 + b_4 \tau^4 + b_5 \tau^5.
\]
The first three coefficients are determined from the initial state:
\[
b_0 = d_0, \quad b_1 = \dot{d}_0, \quad b_2 = \frac{1}{2} \ddot{d}_0.
\]
The terminal state constraints at $\tau = T$ are:

\begin{align*}
d_T &= b_0 + b_1 T + b_2 T^2 + b_3 T^3 + b_4 T^4 + b_5 T^5, \\
\dot{d}_T &= b_1 + 2 b_2 T + 3 b_3 T^2 + 4 b_4 T^3 
            + 5 b_5 T^4, \\
\ddot{d}_T &= 2 b_2 + 6 b_3 T + 12 b_4 T^2 + 20 b_5 T^3.
\end{align*}

These can be written in matrix form to solve for $[b_3, b_4, b_5]$:
\[
\begin{bmatrix}
b_3 \\[1pt] b_4 \\[1pt] b_5
\end{bmatrix}
=
\begin{bmatrix}
T^3 & T^4 & T^5 \\[1pt]
3 T^2 & 4 T^3 & 5 T^4 \\[1pt]
6 T & 12 T^2 & 20 T^3
\end{bmatrix}^{-1}
\begin{bmatrix}
d_T - b_0 - b_1 T - b_2 T^2 \\[1pt]
\dot{d}_T - b_1 - 2 b_2 T \\[1pt]
\ddot{d}_T - 2 b_2
\end{bmatrix}.
\]
This provides the unknown coefficients $[b_3, b_4, b_5]$, completing the lateral quintic trajectory.

\subsection{Scenario analyses}
\label{sec:more_scenario}
\paragraph{Right-turn merging conflict}

\begin{figure}[htbp]
    \centering
    \includegraphics[width=\textwidth]{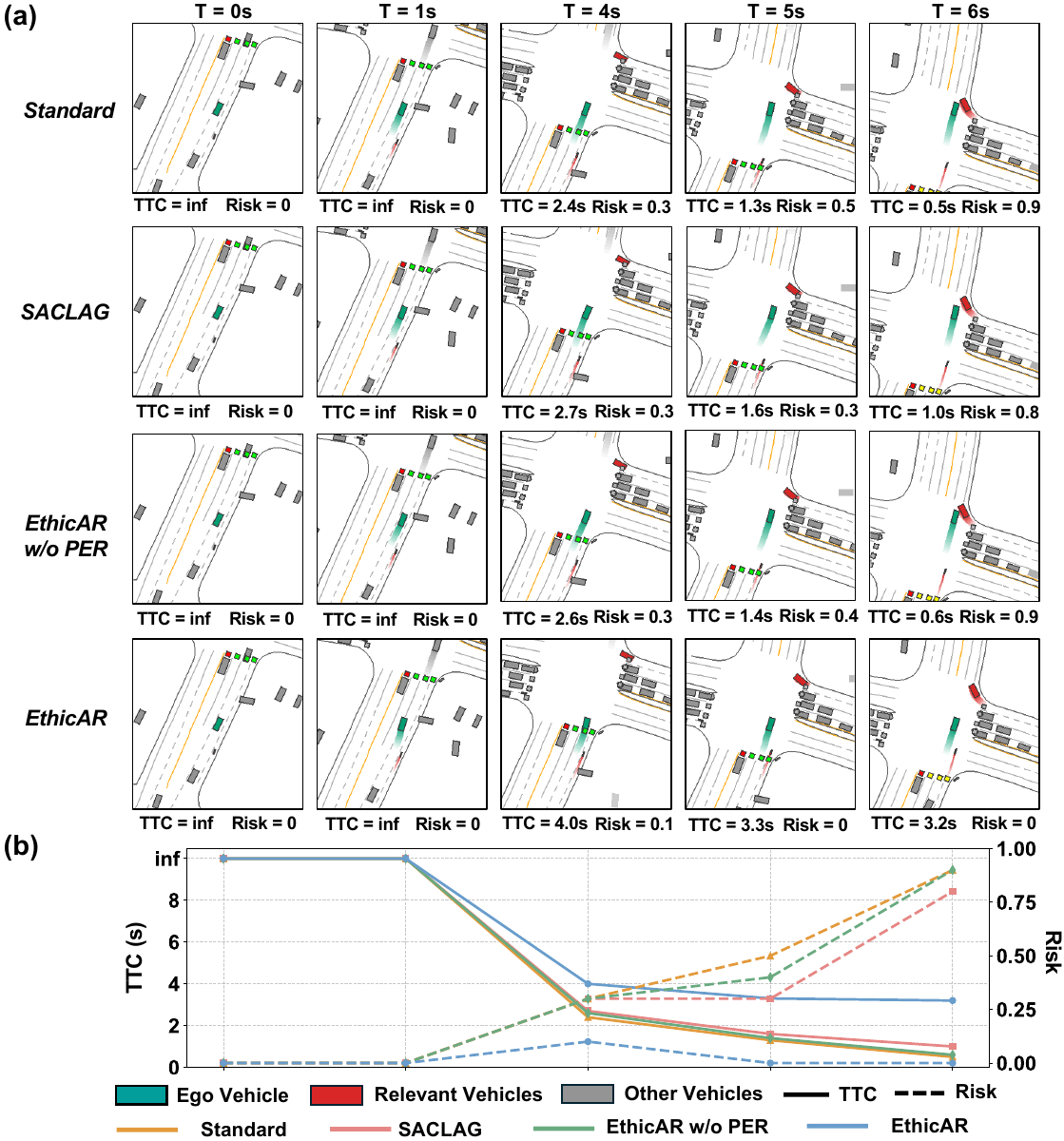}
    \caption{The real-world scenario in which the ego vehicle (green) travels straight alongside a cyclist (red), while a right‑turning vehicle (red) prepares to merge into the main road. (a) are the per-timestep behaviors of EthicAR and other agents, with real-time TTC to the right turn vehicle and maximal risk. (b) is the plot of minimal TTC and maximal risk at the displayed timestep for clear comparison.}
    \label{fig:ttc_risk_ep3}
\end{figure}

This scenario illustrates a typical daily driving conflict: our ego vehicle (green) is traveling along the main road alongside a cyclist on its right, while another vehicle (red), just past the intersection, prepares to turn right and merge onto the main road (see Figure \ref{fig:ttc_risk_ep3}). \par

In this scenario, the ego vehicle has right‑of‑way on the main road, and the right‑turning red vehicle must yield. Three baseline agents, standard mode, SACLAG, and EthicAR w/o PER, maintain a relatively high speed through the intersection. After $T=1\mathrm{s}$, their TTC with the red vehicle decreases, driving up the collision risk. By $T=6\mathrm{s}$, both standard mode and EthicAR w/o PER collide with the red vehicle; SACLAG narrowly avoids impact but still operates in a high‑risk regime. In contrast, the EthicAR agent behaves more cautiously. At $T=1\mathrm{s}$, it drifts slightly left to create space for the cyclist and settles into a comfortable speed. As the red vehicle initiates its turn, the TTC remains relatively constant, and the risks remain low. Even though the ego vehicle legally had right‑of‑way, EthicAR proactively yields to the red vehicle, keeping overall risk at a low level.

\paragraph{Unprotected left-turn conflict}

\begin{figure}[htbp]
    \centering
    \includegraphics[width=\textwidth]{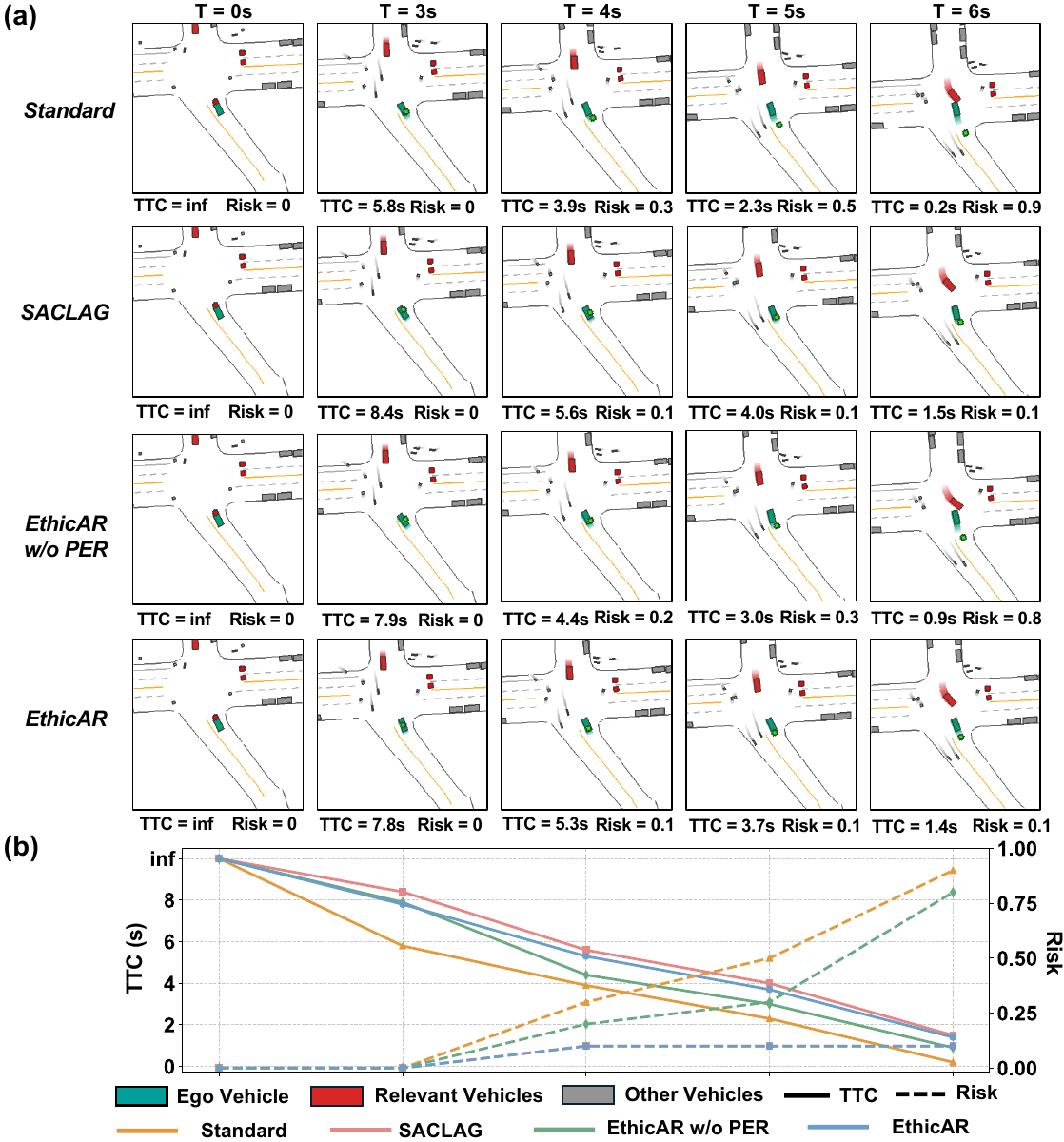}
    \caption{The real-world scenario, going straight ahead with a left turner, with the ego vehicle marked as green driving straight in the intersection, and one vehicle marked in red ready to do a left turn, the background vehicles are marked as gray. (a) are the per-timestep behaviors of EthicAR and other agents, with real-time minimal TTC and maximal risk. (b) is the plot of minimal TTC and maximal risk at the displayed timestep for clear comparison.}
    \label{fig:ttc_risk_ep4}
\end{figure}

Figure \ref{fig:ttc_risk_ep4} depicts a second unprotected left‑turn scenario. Here, the ego vehicle (green) is proceeding straight through the intersection, while another vehicle (red) is poised to turn left. Although the ego vehicle nominally has the right of way, an early‑arriving or aggressively driven red vehicle may still pose a hazard. How, then, should the ego vehicle modulate its behavior to maintain safety when faced with such unpredictable left‑turning traffic? \par

In standard mode, the ego vehicle fails to slow for oncoming traffic at the intersection, ultimately resulting in a collision. The EthicAR w/o PER agent performs slightly better, its minimal deceleration increases the TTC with the red vehicle, but it still adopts a dangerously aggressive approach. In contrast, both SACLAG and the EthicAR agent show the strongest safety performance: they proactively reduce the speed as it nears the intersection, thereby minimizing risk to all road users. By yielding courteously and allowing the red vehicle to complete its left turn first.

\end{document}